\pdfoutput=1
\documentclass[11pt]{article}

\usepackage[]{acl}

\usepackage{times}
\usepackage{latexsym}

\usepackage[T1]{fontenc}
\usepackage[utf8]{inputenc}

\usepackage{microtype}
\usepackage{ulem}
\usepackage{makecell}
\usepackage{amsmath}

\usepackage{booktabs}
\usepackage{enumitem}
\setenumerate[1]{itemsep=0pt,partopsep=0pt,parsep=\parskip,topsep=2pt}
\setitemize[1]{itemsep=0pt,partopsep=0pt,parsep=\parskip,topsep=2pt}
\setdescription{itemsep=0pt,partopsep=0pt,parsep=\parskip,topsep=2pt}

\usepackage{titlesec}
\usepackage{array}
\usepackage{amsmath}
\usepackage{amssymb}
 \usepackage{graphicx}
\DeclareMathOperator*{\argmax}{arg\,max}

\title{FlipDA: Effective and Robust Data Augmentation for Few-Shot Learning}

\author{Jing Zhou \thanks{$\;$ The authors have contributed equally to this work.}$\;^{1}$  ~~Yanan Zheng \footnotemark[1]$\;^{24}$ ~~Jie Tang $^{24}$ ~~Jian Li \thanks{$\;$ Corresponding Authors.}$\;^{1}$ ~~Zhilin Yang \footnotemark[2]$\;^{13}$\\
$^1$Institute for Interdisciplinary Information Sciences (IIIS), Tsinghua University \\
$^2$Department of Computer Science and Technology, Tsinghua University \\
$^3$Shanghai Qi Zhi Institute \\
$^4$Beijing Academy of Artificial Intelligence (BAAI)
}

\begin{document}
\maketitle
\begin{abstract}
Most previous methods for text data augmentation are limited to simple tasks and weak baselines. We explore data augmentation on hard tasks (i.e., few-shot natural language understanding) and strong baselines (i.e., pretrained models with over one billion parameters). Under this setting, we reproduced a large number of previous augmentation methods and found that these methods bring marginal gains at best and sometimes degrade the performance much. To address this challenge, we propose a novel data augmentation method FlipDA that jointly uses a generative model and a classifier to generate label-flipped data. Central to the idea of FlipDA is the discovery that generating label-flipped data is more crucial to the performance than generating label-preserved data. Experiments show that FlipDA achieves a good tradeoff between effectiveness and robustness---it substantially improves many tasks while not negatively affecting the others.\footnote{Code is available at \url{https://github.com/zhouj8553/FlipDA}}
\end{abstract}

\section{Introduction}
Data augmentation is a method to augment the training set by generating new data from the given data. For text data, basic operations including replacement, insertion, deletion, and shuffle have been adopted widely and integrated into a wide range of augmentation frameworks \cite{CharCNN,KNN,UDA,rnn_replacement,EDA}. Generative modeling methods such as back-translation have also been employed to generate augmented samples \cite{FadaeeBM17a_bt,Sennrich_bt}. However, there are two major limitations.
% of the previous studies. 
% First, most of the previous methods are based on weak baselines without using large-scale pretrained language models. Recent work showed that some of the data augmentation methods are less useful when combined with large pretrained models \cite{useless_da}. 
First, some general augmentation methods are based on weak baselines without using large-scale pretrained language models. Recent work showed that some of the data augmentation methods are less useful when combined with large pretrained models \cite{useless_da}. 
% While some effective augmentation methods on large pretrained models \cite{LADA,CRQDA} are not general enough.
Second, most prior studies are carried on simple tasks such as single-sentence classification where it is easier to generate legit augmented samples. For harder tasks such as natural language inference (e.g., telling whether sentence A entails sentence B), it is not clear whether previous methods still help.

% \yanan{shortened: SuperGLUE has been introduced in later section}
This work takes a step further to study data augmentation under strong baselines and hard tasks. 
Our study employs large-scale pretrained language models such as DeBERTa \cite{He2020DeBERTaDB} with over one billion parameters as baselines. 
Moreover, we target a very challenging setting---few-shot natural language understanding (NLU).
Following \cite{Schick2021ItsNJ}, we consider challenging NLU tasks including question answering, textual entailment, coreference resolution, and word sense disambiguation, and use only 32 training examples for each task.
Under this setting, we reproduced several widely-used prior methods for data augmentation. Our experiments lead to two unexpected discoveries: (1) most of prior augmentation methods bring only marginal gains at best and are not effective for most tasks; (2) in many cases, using data augmentation results in instability in performance and even entering a failure mode; i.e., performance may drop by a lot or fluctuate severely depending on which pretrained model is used. The above issues prevent these augmentation methods from practical usage for few-shot learning.

% In this work, we take a step further to study data augmentation under strong baselines and hard tasks. Our study employs large-scale pretrained language models such as DeBERTa \cite{He2020DeBERTaDB} with over one billion parameters as baselines. Moreover, we target a very challenging setting---few-shot natural language understanding (NLU). We consider challenging NLU tasks including question answering, textual entailment, coreference resolution, and word sense disambiguation. We adopt SuperGLUE which was constructed to include some of the most difficult language understanding tasks for current NLP approaches \cite{Wang2019SuperGLUEAS}. Following \cite{Schick2021ItsNJ}, we used only 32 training examples to construct a few-shot setting. Under this setting, we reproduced a large number of widely-used prior methods for data augmentation. Our experiments lead to two unexpected discoveries: (1) most of previous augmentation methods bring only marginal gains at best and are not effective for most tasks; (2) in many cases, using data augmentation results in instability in performance and even entering a failure mode; i.e., performance may drop by a lot or fluctuate severely depending on which pretrained model is used. The above issues prevent these augmentation methods from practical usage for few-shot learning.

We propose a novel method FlipDA that achieves both effectiveness and robustness for hard few-shot tasks.
Preliminary experiments showed that
% During preliminary experiments, we observed that
label-flipped data often largely improve the generalization of pretrained models, compared to augmented data that preserve the original labels. Based on this observation, FlipDA first generates data using word substitution based on a pretrained T5 \cite{Raffel2020ExploringTL} and uses a classifier to select label-flipped data. Experiments demonstrate FlipDA substantially improves performance on many of the hard tasks, outperforming previous augmentation baselines in terms of average performance by a large margin. Moreover, FlipDA is robust across different pretrained models and different tasks, avoiding failure modes.

\section{Related Work}

% \zj{I have combined most of the descriptions of methods, but I don't know how to reduce the analysis statements.}

\paragraph{Data Augmentation.}
% Data augmentation aims to increase the amount of training data, and then utilizes it to improve the performance of the model.
% A wide range of data 
An important type of augmentation methods are based on \textit{word substitution}, such as synonym replacement \cite{CharCNN}, KNN replacement \cite{KNN,KNN2}, Unif replacement \cite{UDA}, TF-IDF replacement \cite{UDA}, Bi-RNN replacement \cite{rnn_replacement}, and other entity replacement methods \cite{RaimanM17_QAentity,Snippext,PHICON} etc. 
% Synonym replacement and KNN replacement are the most popular among them.
% Entity replacement is useful in question answering \cite{RaimanM17_QAentity}, opinion mining \cite{Snippext}, and entity detection \cite{PHICON} tasks.
% EDA \cite{EDA} combines four simple augmentation methods (i.e., synonym replacement, random deletion, random swap, and random insertion) and achieves good results on several single sentence classification tasks.
% Back translation \cite{FadaeeBM17a_bt,Sennrich_bt} is widely used and becoming standard practice in machine translation tasks. Back translation is also adapted to question answering \cite{Yu2018QANetCL} tasks.
EDA \cite{EDA} combines four simple augmentation methods and back translation (BT) \cite{FadaeeBM17a_bt,Sennrich_bt,Yu2018QANetCL} is also widely used.
Unfortunately, EDA and BT are shown to be less useful with large pretrained models \cite{useless_da}.

% Some augmentation methods \cite{ZhangCDL18_mixup,GuoKR20_mixup,Mixtext,LADA,Snippext,feature_space_da} are based on the perturbation in the feature space. 
% Generation based methods \cite{CGBERT_VAE,LiQTC0Y19_VAE,RussoHMZ20_VAE,YooSL19_VAE,SSMBA,CRQDA,HouLCL18_seq2seq_aug} are also proposed.
% In addition, large pretrained models, such as GPT-2, BERT, and Bart have been used for data augmentation \cite{Kumar_bertmlm,LAMBADA,GPT3Mix}. 

Some augmentation methods are based on the \textit{perturbation in the feature space}~\cite{ZhangCDL18_mixup,GuoKR20_mixup,Mixtext,LADA,Snippext,feature_space_da}.  \textit{Generation}~\cite{CGBERT_VAE,LiQTC0Y19_VAE,YooSL19_VAE,SSMBA,CRQDA,HouLCL18_seq2seq_aug} based methods are also proposed for better data diversity.

In addition, large pretrained models have been used for data augmentation. 
\cite{Kumar_bertmlm} utilize large pretrained models, such as GPT-2, BERT, and BART, for conditional data augmentation.
LAMBADA \cite{LAMBADA} finetunes a GPT-2 model with the priming technique to get augmented examples. GPT3Mix \cite{GPT3Mix} uses GPT-3 along with prompting to generate augmented data for classification tasks. 
Our method is similar to this line of work in that we also use pretrained models for generating augmented data. However, there are the following key differences. First, it is challenging for these prior methods to handle long sequences or multiple sentences. In our preliminary experiments, we were not able to use these methods to generate proper data samples (see details in Section \ref{sec:exp}). Second, besides generating augmented samples, we found it crucial to use label-flipped data for augmentation, which is a unique and critical aspect of FlipDA.

\paragraph{Self-training.}
% Self-training \cite{Scudder65ST} labels unlabeled data with a trained model, and then uses the labeled data and unlabeled data jointly to train a new model. This process may repeat for a few iterations. 

Self-training~\cite{Scudder65ST} iteratively augments training data by labeling unlabeled data with a trained model \cite{Yarowsky95_STSenseDisambiguation,Riloff96_ST_extraction_patterns}.
% Self-training~\cite{Scudder65ST} labels unlabeled data with a trained model, and then uses the labeled data and unlabeled data jointly to train a new model. 
% This process may repeat for a few iterations. 
% It is applied in many fields, including sense disambiguation \cite{Yarowsky95_STSenseDisambiguation}, pattern extraction \cite{Riloff96_ST_extraction_patterns}, parsing \cite{McCloskyCJ06_ST_parsing,HuangH09_ST_parsing,ReichartR07a_ST_parsing}, image classification \cite{ST_billion_cls}, neural sequence generation \cite{HeGSR20_ST_NMT}, speech recognition \cite{Kahn0H20_ST_speech}, etc.
Knowledge distillation and pseudo-labeling are special forms of self-training \cite{Hinton2015DistillingTK,Lee2013_pseudo,ReedLASER14_pseudo}.
Strong data augmentation \cite{ZophGLCLC020_rethinking_PT_ST}, equal-or-larger model \cite{XieLHL20_noisy_student}, additional noise \cite{XieLHL20_noisy_student, HeGSR20_ST_NMT}, and feedback of the student's performance \cite{ST_meta} are helpful for self-training.
% \cite{ZophGLCLC020_rethinking_PT_ST} observed that different from pre-training, self-training is helpful with strong data augmentation. Noisy Student \cite{XieLHL20_noisy_student} obtained good results on ImageNet by using an equal-or-larger student model and adding noise to the student. \cite{HeGSR20_ST_NMT} concluded that dropout is important for self-training in sequence generation tasks. \cite{ST_meta} proposed to update the parameters of the teacher model using the feedback of the student's performance on labeled data.

Self-training bears similarity to the second phase of FlipDA where a teacher model is used to filter samples. 
Different from self-training, FlipDA 
% discovers and leverages the usefulness 
leverages the advantages of label flipping to improve performance and does not rely on unlabeled data. 

\paragraph{Label Flipping.}
Our manual label flipping augmentation procedure is analogous to \cite{KaushikHL20_manual_flip} and \cite{ABBBCDDEGGH20_manual_contrast}. 
\citet{KaushikHL20_manual_flip} aimed to mitigate the effects of learning spurious features. 
\citet{ABBBCDDEGGH20_manual_contrast} targeted reducing systematic gaps in the dataset. 
In contrast, we target improving few-shot generalization. Moreover, we measure the performance on an existing i.i.d. test set while \citet{KaushikHL20_manual_flip} and \citet{ABBBCDDEGGH20_manual_contrast} created more challenging test sets. Most importantly, we propose an automatic method of label flipping, going beyond manual efforts.

\paragraph{Contrastive Learning.}
% Few-shot learning systems are prone to overfitting for their lack of samples. 
% The core idea of FlipDA is to provide intermediate supervision about what causes a label difference to improve generalization.
% So the best method is to 'tell' it what to learn. If we show the model some similar samples, the model may learn what representations are similar, which could have been learned during pre-training. While if we give it two samples with similar expressions but different labels, it will try to learn the difference.
% In this sense, there is a connection between FlipDA and contrastive learning (CL)~\cite{He0WXG20_MOCO,ChenK0H20_SIMCLR}.
FlipDA is connected to contrastive learning (CL)~\cite{He0WXG20_MOCO,ChenK0H20_SIMCLR} in that they both improve generalization by considering label differences.
CL uses data augmentation to generate positive instances and uses samples existing in the dataset as negative samples, while FlipDA shows that negative samples can be automatically generated. 
While previous work on CL focuses on training with large datasets, our experiments show that augmenting a small dataset can improve few-shot generalization.
% \yanan{recheck the deletion}
It could be intriguing to see whether such a connection might lead to advances in both fields, e.g., generating negative samples for large-scale contrastive pretraining.

% \section{Data Augmentation for Few-Shot Learning}

% \vspace{-5pt}
\section{Few-Shot Data Augmentation}
% \vspace{-10pt}

\subsection{Setting}\label{sec:setting}

% In this work, we consider data augmentation on hard tasks and strong baselines.

% \yanan{recheck the following sec}
\paragraph{Few-Shot NLU Tasks.}
This work considers a collection of ``difficult'' NLU tasks from SuperGLUE~\cite{wang2019superglue} that require in-depth understanding of the input in order to obtain high performance, including coreference resolution \cite{Levesque2011TheWS}, causal reasoning \cite{Gordon2012SemEval2012T7}, textual entailment \cite{Marneffe2019TheCI,Dagan2005ThePR}, word sense disambiguation \cite{Pilehvar2019WiCTW}, and question answering \cite{Clark2019BoolQET,Khashabi2018LookingBT,Zhang2018ReCoRDBT}.
Following \citet{Schick2021ItsNJ}, we used only 32 training examples to construct a few-shot setting to further increase the difficulty.

% Natural language understanding is a collection of tasks that require in-depth understanding of the input in order to obtain high performance. NLU tasks range from coreference resolution \cite{Levesque2011TheWS}, causal reasoning \cite{Gordon2012SemEval2012T7}, textual entailment \cite{Marneffe2019TheCI,Dagan2005ThePR}, and word sense disambiguation \cite{Pilehvar2019WiCTW} to question answering \cite{Clark2019BoolQET,Khashabi2018LookingBT,Zhang2018ReCoRDBT}. These tasks are usually formulated as mapping a sentence or multiple sentences to a certain label. For systematic evaluation, we adopt SuperGLUE which contains a set of NLU tasks and is designed to benchmark progress on ``difficult'' language understanding capabilities for current NLP approaches \cite{wang2019superglue}. Following \citet{Schick2021ItsNJ}, we used only 32 training examples to construct a few-shot setting to further increase the difficulty.

% We target few-shot learning on SuperGLUE \cite{wang2019superglue} following the settings of \cite{Schick2021ItsNJ}. SuperGLUE contains a set of natural language understanding tasks and is designed to benchmark progress on ``difficult'' language understanding capabilities for current NLP approaches. Following \cite{Schick2021ItsNJ}, we used only 32 training examples to construct a few-shot setting. 

\paragraph{Large-Scale Pretrained Models.} Our setting assumes a large-scale pretrained language model \cite{Devlin2019BERTPO,Lan2020ALBERTAL,He2020DeBERTaDB} is available and few-shot learning is performed based on the pretrained model. This setting is crucial since previous studies found that using a strong pretrained model as the baseline eliminates the benefits of data augmentation \cite{useless_da} while large pretrained models are becoming more and more available. Our main result is based on DeBERTa \cite{He2020DeBERTaDB} with over one billion parameters. We also provide results with ALBERT which has fewer parameters \cite{Lan2020ALBERTAL}.

% \yanan{The following Sec could be summarized into one simple sentence.}
% \yanan{recheck the deletion.}
\paragraph{Preliminary Experiments with Prior Methods.} 
% Our preliminary experiments with a large number of previous methods lead to a conclusion that there is not an effective and robust method available for this hard setting. 
% With the previous methods, the gains are limited while it is possible to enter a failure mode with substantial performance drop. 
% More details will be discussed in Section \ref{sec:exp}. 
% We will discuss how we tackle this challenge by proposing a novel data augmentation method FlipDA in the following sections.
Our preliminary experiments with a large number of previous methods (in Section \ref{sec:exp}) lead to a conclusion that there is not an effective and robust method available for this hard setting. 
% With the previous methods, the gains are limited while it is possible to enter a failure mode with substantial performance drop. 
% More details will be discussed in Section \ref{sec:exp}.
We will discuss how we tackle this challenge by proposing a novel data augmentation method FlipDA in later sections.

% \subsection{Preserving Labels is Challenging}
% The first is that it is harder to keep its label, compared to sentiment classification tasks.
% Almost all tasks in FewGLUE are two-sentence tasks. If we augment each sentence alone, we will face a dilemma: each sentence seems to get a perfect augment, but their relationship is changed.

% Let's look at an example.
% previous example: {"premise": "This case of rabies in western Newfoundland is the first case confirmed on the island since 1989.", "hypothesis": "A case of rabies was confirmed.", "label": "entailment"}
% augmented: "premise": "This case of cholera in western Newfoundland is the first case confirmed on the island since 1989.", "hypothesis": "A case of rabies was confirmed."
% Here, both rabies and cholera are both the name of the infectious disease, they have the same sentiment, and similar representations, but their label is different.

% From this example, we can see that, if we want to get a good augmentation result, we must consider the relationship between two sentences.
% Fortunately, predicting the prompt is a good way to solve it. Prompt is a  natural bridge connecting two sentences

\subsection{Desiderata: Effectiveness and Robustness}

We propose key desiderata for data augmentation methods under the setting of few-shot learning.
\begin{enumerate}
    \item\textbf{Effectiveness.} A data augmentation method should be able to improve performance on certain tasks in a significant manner.
    \item\textbf{Robustness.} A data augmentation method should not suffer from a failure mode in all cases. Failure modes are common for few-shot learning where some minor changes might cause substantial performance drop. We argue this should be used as a key evaluation metric. 
    %We mainly consider two types of robustness in this work:
    We consider two types of robustness:
    (1) robustness w.r.t. different base pretrained models and (2) robustness w.r.t. various tasks.
\end{enumerate}

% \yanan{recheck the deletion.}
% In other words, we want a data augmentation method that improves some tasks while not hurting the others, so as to achieve strong performance in terms of both effectiveness and robustness.

\subsection{Effectiveness: Manual Label Flipping Improves Performance} \label{sec:eff}

% We only conduct experiments on a subset of dataset because manual augmentation is time costing, and we think it is enough to show the effectiveness of flipping label. We will release our manually augmented dataset for reference.
% \textbf{``premise''}: ``Humic acids are complex organic molecules formed by the breakdown of organic matter in the soil. They are not considered to be fertilizers, but soil enhancers and improvers.''\\
% \textbf{``hypothesis''}: ``Organic fertilizers are used as soil enhancers.'' (``not entailment'')\\
% \color[HTML]{3166FF} \textbf{``hypothesis''}: ``humic acids are used as soil enhancers.'' (``entailment'')

% Since previous methods are not sufficiently effective and robust in our preliminary experiments (see Tables \ref{tab:albert} and \ref{tab:deberta} in Section \ref{sec:exp} for more details), we use manual augmentation to investigate what kind of augmented data is beneficial for large pretrained models in the few-shot setting. We mainly study two types of data augmentation---one that preserves the labels and the other that flips the labels. Since manual augmentation is time consuming, we select a subset of representative SuperGLUE tasks in this study.

Since previous methods are not sufficiently effective and robust in our preliminary experiments (see Tables \ref{tab:albert} and \ref{tab:deberta} in Section \ref{sec:exp} for details), we use manual augmentation to investigate what kind of augmented data is beneficial for large pretrained models in the few-shot setting. We mainly study two types of data augmentation---one that preserves the labels and the other that flips the labels. Since manual augmentation is time consuming, we select a subset of representative SuperGLUE tasks here.

To augment label-flipped data, the following principle is applied---making minimal changes to the original text sample to alter the label. Augmentation includes word addition, deletion, and substitution. To augment label-preserved data, we substitute some of the words with semantically similar words but make sure that the label is unchanged.

\begin{table}[h]
  \caption{\small{Manual data augmentation results. We manually write augmented examples that preserve or flip the label. Flipping the labels substantially improves performance on CB, RTE and WSC by up to 10 points, while preserving the labels only has minor gains.}}
  \label{baseline}
  \centering
%   \scriptsize
\small
%   \vspace{-5pt}
  \begin{tabular}{lccc}
    \toprule[1pt] 
    % Task & Baseline & Paraphrase & Position Changed & Label Flip \\
    % \hline
    % BoolQ & 78.21$\pm$0.27 & 78.55$\pm$0.49 & 78.50$\pm$0.34, 78.31$\pm$0.16 & 77.68$\pm$0.08\\
    % CB & 81.55/72.16$\pm$4.12/7.02 & 82.14/77.07 $\pm$ 3.57/4.91 & 85.71/81.17 $\pm$ 1.79/2.16 & 91.07/88.14 $\pm$ 3.09/3.93\\
    % COPA & 90.33$\pm$1.15 & 91.33$\pm$0.58 & 91.33$\pm$0.58 & 90.33$\pm$0.58\\
    % RTE & 68.11$\pm$3.28 & 67.63$\pm$2.61 & 70.04$\pm$3.28, 67.15$\pm$1.25 & 76.05 $\pm$ 0.75 \\
    % WiC & 50.52$\pm$1.18 & 51.88$\pm$0.54& 49.95$\pm$0.48 & -\\
    % WSC & 79.49$\pm$2.22 & 78.53$\pm$2.78 & 83.97$\pm$2.22 & 85.58$\pm$0.96\\
    % MultiRC & 37.78/76.66$\pm$0.84/0.19 & 37.08/76.93 $\pm$ 0.84/0.46 & 37.63/76.92 $\pm$ 0.80/0.41 & -\\
    % Aug Method & Acc. & Acc./F1 & Acc. & Acc. & Acc. & EM/F1a \\
    
    Tasks & No DA & Preserves & Flips \\
    \midrule
    % \hline
    BoolQ & 78.21$\pm$0.27 & \textbf{78.55}$\pm$0.49 & 77.68$\pm$0.08\\
    CB-Acc & 81.55$\pm$4.12 & 82.14$\pm$3.57 & \textbf{91.07}$\pm$3.09\\
    CB-F1 & 72.16$\pm$7.02 & 77.07$\pm$4.91 & \textbf{88.14}$\pm$3.93 \\
    COPA & 90.33$\pm$1.15 & \textbf{91.33}$\pm$0.58 & 90.33$\pm$0.58\\
    RTE & 68.11$\pm$3.28 & 67.63$\pm$2.61 & \textbf{76.05}$\pm$0.75 \\
    WSC & 79.49$\pm$2.22 & 78.53$\pm$2.78 & \textbf{85.58}$\pm$0.96\\
    \bottomrule[1pt]
  \end{tabular}
\end{table}

\begin{table*}[h]
  \caption{\small{Label-flipped examples from manual augmentation. The augmentation principle is to make minimal changes that are sufficient to alter the labels. Black denotes original examples, and blue denotes augmented examples. The second task WSC is coreference resolution, which is to extract the referred entitiy from the text. In this case, ``label'' is defined as the referred entity (denoted in red), and label flipping is defined as modifying the entity.}}
%   \vspace{-5pt}
  \label{tab:case-manual}
  \centering
  \small
  \begin{tabular}{p{10pt}<{\centering} l}
    \toprule[1pt]
    % \Xhline{1pt}
        \begin{tabular}[c]{@{}c@{}}RTE\end{tabular} &\begin{tabular}[c]{p{0.85\textwidth}}
    % \textbf{Original Example}: \\
    ~\textbf{Premise:} This case of rabies in western Newfoundland is the first case confirmed on the island since 1989.\\
    ~\textbf{Hypothesis:} A case of rabies was confirmed.~~~~~~~~\textbf{Entailment:} True\\
    ~{\color[HTML]{3166FF}\textbf{Hypothesis:} A case of smallpox was confirmed.~~~~~~~~\textbf{Entailment:} False} \\
    \end{tabular}\\
    % \hline
    % \begin{tabular}[c]{@{}c@{}}CB\end{tabular} &\begin{tabular}[c]{p{0.85\textwidth}}
    % \textbf{``hypothesis''}: ``even someone as sensible as Miss van Williamsburgh would try to make a play of this sort'' \\
    % \textbf{``premise''}: ````For such a person, finding a protector might not be so difficult, even in Edinburgh.'' Jean smiled. He might have known that even someone as sensible as Miss van Wiliamsburgh would try to make a play of this sort.'' (``entailment'')\\
    % \color[HTML]{3166FF}\textbf{``premise''}:````For such a person, finding a protector might not be so difficult, even in Edinburgh.'' Jean smiled. Do you think that even someone as sensible as Miss van Wiliamsburgh would try to make a play of this sort?'' (``neutral'')\\
    % \end{tabular} \\
    \midrule
    % \hline
    \begin{tabular}[c]{@{}c@{}}WSC\end{tabular} &\begin{tabular}[c]{p{0.85\textwidth}}
    ~\textbf{Text:} The city councilmen refused {\color{red}the demonstrators} a permit because \underline{they} advocated violence. \\
     ~{\color[HTML]{3166FF}\textbf{Text:} The city councilmen refused {\color{red}the criminals} a permit because \underline{they} advocated violence.}\\
    \end{tabular} \\
    \bottomrule[1pt]
    % \Xhline{1pt}
    \end{tabular} \\
% \vspace{-10pt}
\end{table*}

Results are shown in Table \ref{baseline}.\footnote{For each original example, we produce one augmented example for each type. The augmented data and the original data are combined for training. Following \citet{Schick2021ItsNJ}, we train each pattern with three seeds and ensemble these (pattern, seed) pairs. We repeat this ensemble process 3 times and report their mean and standard deviation.} Flipping labels substantially improves performance on three of the tasks by up to 10 points, while preserving the labels only has minor gains. In contrast, many 
prior methods on data augmentation focus on creating data examples that are assumed to have the same labels as the original ones. This might explain why previous augmentation methods are not sufficiently effective for the few-shot setting.
% Some of the label-flipped augmented examples are shown in Table \ref{tab:case-manual}. We conjecture that label flipping augmentation provides useful information about the important components in a sentence that determine the label. In other words, augmented samples provide intermediate supervision that explains the predictions, which improves generalization in a few-shot setting.
Some of the label-flipped augmented examples are shown in Table \ref{tab:case-manual}. We conjecture that label flipping augmentation provides useful information about the important components in a sentence that determine the label.
% In other words, augmented samples provide intermediate supervision that explains the predictions, which improves generalization in a few-shot setting.
In other words, augmented samples provide intermediate supervision that explains the predictions, improving generalization in a few-shot setting.

There is a caveat about this manual augmentation experiment. 
% Although we follow a certain principle (i.e., making minimal changes to alter the label) and pay much attention to the augmentation quality, the manual augmentation procedure is inevitably subjective and hard to reproduce.
Although we follow certain principles and pay much attention to the augmentation quality, the manual augmentation procedure is inevitably subjective and hard to reproduce.
For reference, we will make our manually augmented dataset publicly available. More importantly, we will design an automatic method (FlipDA) in the following sections for objective evaluation and reproducibility. 
% That said, the findings in this section motivate the core idea of FlipDA.

% What's more, manual augmentation is time costing and can be greatly influenced by the writer, so an automatic label-flipped data augmentation method is necessary.

\subsection{Robustness: What Contribute to Failure Modes?} \label{sec:robust}
% \zj{subsection too long, shortened}
% We also use preliminary experiments to analyze why augmentation methods usually suffer from failure modes. Most augmentation methods are based on a label preserving assumption that the newly generated data samples have the same labels as the original ones. However, it is challenging for automatic methods to always generate samples that preserve the labels in a hard NLU setting.

We also analyze why augmentation methods usually suffer from failure modes. 
Most augmentation methods are based on a label preserving assumption, while it is challenging for automatic methods to always generate label-preserved samples.
We first examine the samples generated by prior automatic methods EDA~\cite{EDA} and KNN~\cite{KNN} in Table~\ref{tab:bad-cases}. 
In the first example, the keyword ``rabies'' is deleted, which not only results in a grammatically incorrect expression but also eliminates the key information to support the hypothesis.
In the second example, the ``Lake Titicaca'' is replaced by ``Lake Havasu'', which results in a label change from entailment to non-entailment. 
If a model is trained on these noisy augmented data with the label preserving assumption, performance degradation is expected.

% To further verify the cause of failure modes, 
We further experimented with EDA~\cite{EDA} on the RTE task~\cite{Dagan2005ThePR} to verify the cause of failure modes.
Using EDA decreases the performance by a few percentage points with both ALBERT and DeBERTa, entering a failure mode. 
% To analyze the reason, 
We identified two types of noise in the augmented samples: (1) grammatical errors that lead to the difficulty of understanding and (2) modification of key information that alters the labels. We experimented with (1) replacing these noisy samples with the original ones and (2) correcting the labels of the noisy samples.~\footnote{For label correction, if a sample has severe grammatical mistakes and is not understandable by human, we always mark it as ``not entailment''. This is related to an interesting phenomenon that label flipping is usually asymmetric for NLU tasks. We will discuss more of the phenomenon in Section \ref{sec:trans}.}
% As Table~\ref{tab:eda_rte} shows, both operations of replacing and correcting noisy samples largely improve performance to prevent the failure mode. Moreover, correcting the labels brings large gains and results in favorable performance compared to the baseline without data augmentation. This indicates that the label-preserving assumption and grammatical errors contribute to failure modes, and label flipping tends to alleviate the issue.
As Table~\ref{tab:eda_rte} shows, both replacing and correcting noisy samples largely improve performance to prevent the failure mode. Moreover, correcting the labels brings large gains, indicating label flipping tends to alleviate the issue.

% is prone to error, and correcting this type of error is key to a robust augmentation method.

% wrong-labeled augmented samples and experimented with (1) replacing these wrong-labeled samples with the original ones and (2) correcting the label. 

% Examples of data augmented by EDA on the RTE task are shown in Table \ref{tab:bad-cases}. Since EDA does not employ language modeling, more often than not, it produces sentences with substantial grammatical errors and the label is changed from ``entailment'' to ``not entailment''. This phenomenon of asymmetric label transformation is further studied in Section \ref{sec:trans}. This indicates that using an augmentation method with fewer grammatical errors might also be essential for preventing failure modes on NLU.

% Through the experiments, we identify two major factors that contribute to the failure modes of previous data augmentation methods: (1) the label preserving assumption and (2) substantial grammatical errors. Again, 

% Similar to experiments in Section \ref{sec:eff}, experiments in this section involve subjective factors like human judgement. 
To reiterate, these experiments involve subjective factors and are merely meant to show the intuition of FlipDA, rather than proving its superiority.

% \todo{add a table and rewrite} We take a look at the samples augmented by EDA \cite{EDA} on RTE, where the performance drops from 61.40 to 58.33 with ALBERT (see Table \ref{tab:albert}) and from 81.95 to 77.38 (-4.57) on DeBERTa (see Table \ref{tab:deberta}). We urged to find out why its performance drops so much with this data augmentation method. According to our observation, the augmented samples face two key issues: wrong labels and illegal grammar. To explore the effect of wrong labels, we conduct two experiments. The first is replacing the wrong-label augmented samples with corresponding original samples, and the second is flipping the labels of the wrong-label augmented samples. We find that about 16.32\% augmented samples are with wrong labels, and some cases are in Table \ref{tab:bad-cases}. On DeBERTa, the first experiment achieves a score of 80.75, and the second achieve a score of 83.39. On ALBERT, the first experiment achieves a score of 59.39, and the second achieve a score of 61.07. We can see that augmented samples with wrong labels can explain most of the performance degradation, and correcting the label can contribute to a better performance, which is the same as what we described in Section \ref{sec:eff}.
\begin{table}
\small
\caption{\small{Performance of correcting the wrong-labeled augmented data by EDA on RTE. W-Del denotes replacing the wrong-labeled augmented samples with corresponding original samples, and W-Flip denotes flipping the labels of the wrong-labeled augmented samples to be the correct ones. The results show that in this case data augmentation with the label-preserving assumption substantially contributes to performance drop.}}
% \vspace{-5pt}
\label{tab:eda_rte}
\centering
\begin{tabular}{lllllllll}
\toprule[1pt]
% \Xhline{0.75pt}
 & No DA & EDA & W-Del & W-Flip \\
%  \hline
\midrule
 ALBERT & 61.40 & 58.33 & 59.39 & 61.07 \\
 DeBERTa & 81.95 & 77.38 & 80.75 & 83.39 \\
% \Xhline{0.75pt}
\bottomrule[1pt]
\end{tabular}
% \vspace{-5pt}
\end{table}

\begin{table*}[h]
  \caption{\small{Augmented example with wrong labels. The first is by EDA, and the second is by KNN. Black denotes original examples, blue denotes augmented examples and red denotes key entity. The phenomenon of asymmetric label transformation (e.g., flipping from ``entailment'' to ``not entailment'' is more common) is further studied in Section \ref{sec:trans}.}}
%   \vspace{-5pt}
  \label{tab:bad-cases}
  \centering
  \small
  \begin{tabular}{c}
  \toprule[1pt]
    %     \begin{tabular}[c]{@{}c@{}}Bad Case \end{tabular}                     & \begin{tabular}[c]{p{0.7\textwidth}}
    % \textbf{Original Example}: \\
    
    {\begin{tabular}[c]{p{0.95\textwidth}}
    \textbf{Premise:} This case of {\color{red}{rabies}} in western Newfoundland is the first case confirmed on the island since 1989.\\
    \textbf{Hypothesis:} A case of rabies was confirmed.~~~~~~~~\textbf{Entailment:} True\\
    
    \end{tabular}}
     \\
    % \textbf{Augmented Example}:\\
    {\color[HTML]{3166FF}
    \begin{tabular}[c]{p{0.95\textwidth}}
    \textbf{Premise:} this case of in western newfoundland is the first case confirmed on the island since 1989.\\
    \textbf{Hypothesis:} a case of rabies was confirmed.~~~~~~~~\textbf{Entailment:} False\\
    \end{tabular}}\\
    % \end{tabular} \\
    \hline
    % {\begin{tabular}[c]{p{0.95\textwidth}}
    % \textbf{``premise''}: ``officials claimed they were backed by influential members of the santa cruz business community of croatian descent . the security vice - minister , marcos farfan , said that investigators have surveillance photographs of mr farrell at various public events hosted by mr morales , includes a peasant rally near santa cruz and a visit to naval installations on lake titicaca , mr farfan said that mr dwyer was \" following \" mr sanchez and other officials as part of the preparations for of me assassination plot \" . he added that police experts are analysing contents reportedly found in computers taken from the rooms in which the men were killed .''\\
    % \textbf{``hypothesis''}: ``Lake Titicaca has a naval installation.'' (``entailment'')\\
    {\begin{tabular}[c]{p{0.95\textwidth}}
    \textbf{Premise:} ... including a peasant rally near Santa Cruz and a visit to naval installations on Lake Titicaca ...\\
    \textbf{Hypothesis:} {\color{red}Lake Titicaca} has a naval installation.~~~~~~~~\textbf{Entailment:} True\\
    \end{tabular}}
    \\
    % \textbf{Augmented Example}:\\
    {\color[HTML]{3166FF}
    \begin{tabular}[c]{p{0.95\textwidth}}
    \textbf{Premise:} ... includes a peasant rally near santa cruz and a visit to naval installations on lake titicaca ...\\
    \textbf{Hypothesis:} {\color{red}lake havasu} has a naval installation .~~~~~~~~\textbf{Entailment:} False\\
    \end{tabular}}\\
    % \end{tabular} \\
    % \hline
    %         % \begin{tabular}[c]{@{}c@{}}Manual DA\end{tabular}                     & \begin{tabular}[c]{p{0.7\textwidth}}
    % % \textbf{Original Example}: \\
    % {\begin{tabular}[c]{p{0.95\textwidth}}
    % \textbf{``premise''}: ``Chinese Premier, Zhu Rongji, and German Chancellor, Gerhard Schroeder, opened the world's first commercial magnetic levitation (maglev) train yesterday, with both sides having much to gain from its success.''\\
    % \textbf{``hypothesis''}: ``Both Mr. Schroeder and Mr. Zhu have much to gain if maglev succeeds.'' (``entailment'')\\
    
    % \end{tabular}}
    %  \\
    % % \textbf{Augmented Example}:\\
    % {\color[HTML]{3166FF}
    % \begin{tabular}[c]{p{0.95\textwidth}}
    % \textbf{``premise''}: ``success. premier, zhu rongji, and german chancellor, gerhard sides opened the worlds first commercial magnetic levitation (maglev) train yesterday, with both schroeder, having much to gain from its chinese''\\
    % \textbf{``hypothesis''}: ``both mr. zhu and mr. schroeder have much to gain if maglev succeeds.''  (``not entailment'')\\
    % \end{tabular}}\\
    % % \end{tabular} \\
    % \hline
    \bottomrule[1pt]
    \end{tabular} \\
% \vspace{-5pt}
\end{table*}

% \vspace{-1pt}
\subsection{FlipDA: Automatic Label Flipping} \label{sec:flipda}

Observations in Sections \ref{sec:eff} and \ref{sec:robust} show that label-flipping could benefit few-shot NLU in both effectiveness and robustness.
Reducing grammatical errors is also key to preventing failure modes.
This motivates our development of FlipDA that automatically generates and selects label-flipped data without label-preserving assumption.

% The above observations in Sections \ref{sec:eff} and \ref{sec:robust} show that a label flipping data augmentation method without the label-preserving assumption might be beneficial for few-shot NLU, in terms of both effectiveness and robustness. Moreover, reducing grammatical errors is also key to preventing failure modes. This motivates our development of the new method FlipDA that automatically generates and selects label-flipped data, which does not rely on the label-preserving assumption.

\begin{figure*}[h]
  \centering {\includegraphics[width=1.0\linewidth] {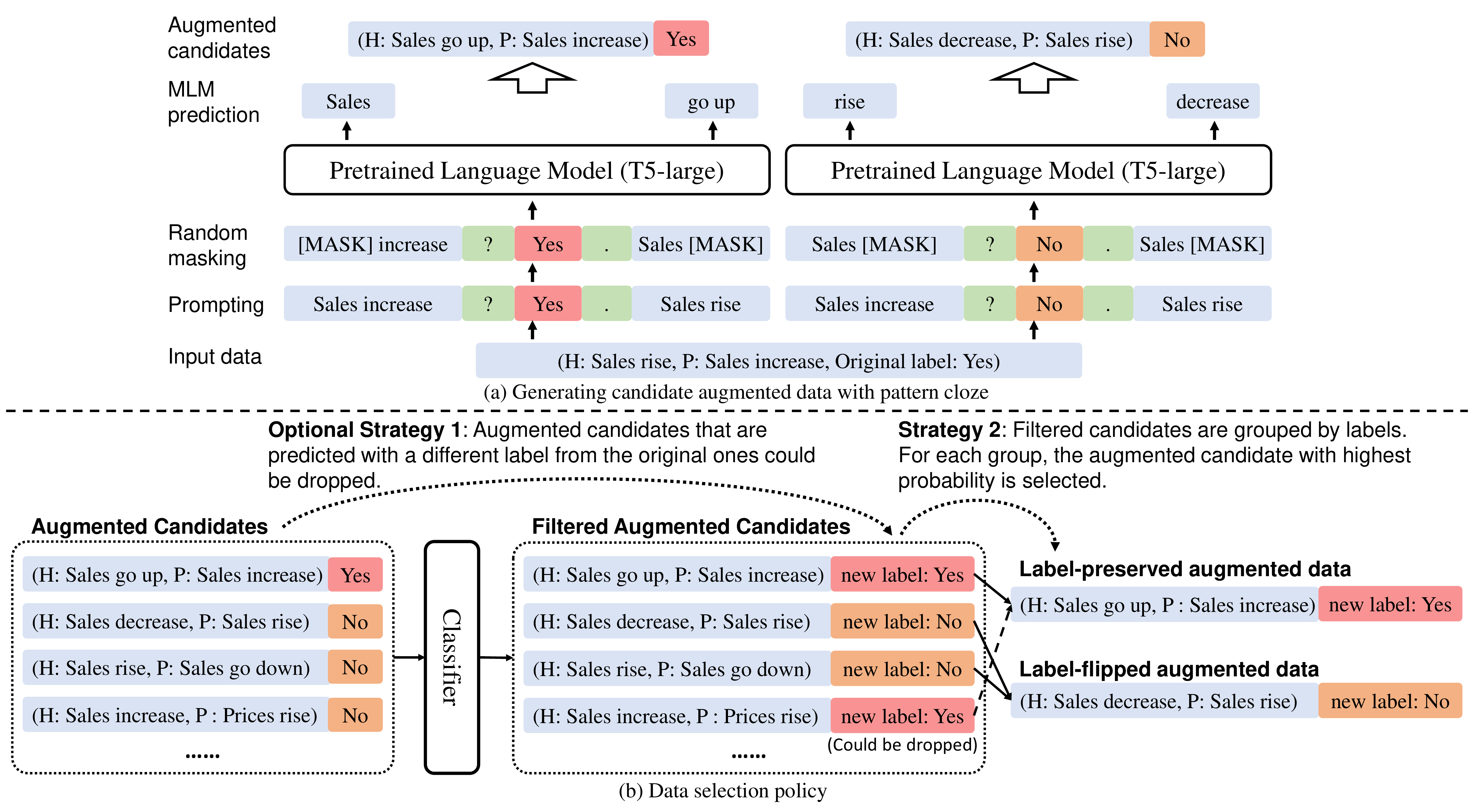}}
%   \caption{Training process of FlipDA. We first train a classifier with standard PET. And then, we generate augmented data with both kept/flipped labels. Thirdly, we utilize the trained classifier to filter the augmented data. Finally, we retrain the model with the original data and the selected augmented data and get a new model.}
  \caption{\small{An illustration of (a) our prompt-based augmentation algorithm for both preserved/flipped labeled data, and (b) our data selection policy. Whether to use Strategy 1 depends on the relative power of the augmentation model and the classification model. If the augmentation model is accurate enough, drop the candidates with inconsistent labels, and otherwise, keep it.}}
  \label{model_figure}
%  \vspace{-10pt}
\end{figure*}

% FlipDA mainly consists of the following steps, as shown in Figure \ref{model_figure}:
% \begin{enumerate}
%     \item Train a classifier (e.g., finetuning a pretrained model) without data augmentation
%     \item Generate augmented samples
%     \item Use the classifier to select generated samples with largest label-flipping probabilities
%     \item Retrain the classifier with additional label-flipped samples
% \end{enumerate}

FlipDA consists of 4 steps as shown in Figure \ref{model_figure}:
%FlipDA mainly consists of the following steps, as shown in Figure \ref{model_figure}:
\begin{enumerate}
    \item Train a classifier (e.g., finetuning a pretrained model) without data augmentation.
    \item Generate label-preserved and label-flipped augmented samples.
    \item Use the classifier to select generated samples with largest probabilities for each label.
    \item Retrain the classifier with the original samples and the additional augmented samples.
\end{enumerate}

Formally, given a few-shot training set $\{(x_i, y_i)\}_i$ where $x_i$ is text (possibly a set of text pieces or a single piece) and $y_i \in \mathcal{Y}$ is a label. We finetune a pretrained model $f$ to fit the conditional probability for classification $f(x, y) = \hat{p}(y | x)$. 
%For example, the model $f$ can be a pretrained BERT \cite{Devlin2019BERTPO} or its variants \cite{Lan2020ALBERTAL,He2020DeBERTaDB}.
In the second step, we generate augmented samples from the original ones. For each training sample $x_i$, we generate a set of augmented samples $S_i = \{\Tilde{x}_{i, 1}, \Tilde{x}_{i, 2}, \cdots\}$. In our implementation, we first use a cloze pattern \cite{Schick2021ItsNJ} to combine both $x$ and $y$ into a single sequence, and then randomly mask a fixed percentage of the input tokens. This is followed by employing a pretrained T5 model \cite{Raffel2020ExploringTL} to fill the blanks to form a new sample $x'$ (see Appendix \ref{apdx:cloze} for more details). We find it beneficial to remove the sample if T5 does not predict $y$ given $x'$. Note that using T5 to generate augmented samples does introduce additional knowledge and reduce grammatical errors, but naively using T5 for augmentation without label flipping and selection does not work well (see ablation study in Section \ref{sec:exp}). After generating the augmented samples, we use the classifier $f$ for scoring. 
% Specifically, given a generated set of examples $S_i$ from the example $(x_i, y_i)$, we select the example with the highest probability of label flipping, i.e.,
% \[
% x', y' = \argmax_{x \in S_i, y \not= y_i} \hat{p}(y | x)
% \]
% where $x'$ is a sample in the generated set, $y'$ is the flipped label, and the estimated probability $\hat{p}(y' | x')$ scored by the model $f$ is the largest in $S_i$. After selecting the label-flipped example $(x', y')$, if $y'$ is the predicted label of $x'$ (i.e., $y' = \argmax_y \hat{p}(y | x')$), we add $(x', y')$ to the augmented training set; otherwise we discard this example. In other words, we only add an example into the training set if the model $f$ considers the flipped label to be correct. 
% The two implementation methods are not the same (even the same in the binary classification tasks and most cases in the multi-class classification tasks).
Specifically, let $S_i$ be a set of augmented samples generated from the original sample $(x_i, y_i)$. For each label $y' \not= y_i$, we construct a set
\[
\setlength{\abovedisplayskip}{0pt}
\setlength{\belowdisplayskip}{0pt}
S_{i, y'} = \{x | x \in S_i \text{~and~} y' = \argmax_y \hat{p}(y | x)\}\]
which contains all augmented samples with $y'$ being highest-probability class. Given the set $S_{i, y'}$, we select the sample with the highest predicted probability
\[
\setlength{\abovedisplayskip}{0pt}
\setlength{\belowdisplayskip}{0pt}
x', y' = \argmax_{x \in S_{i, y'}, y=y'} \hat{p}(y | x)
\]
where $x'$ is a sample in the generated set, $y'$ is the flipped label, and the estimated probability $\hat{p}(y' | x')$ scored by the model $f$ is the largest in $S_{i, y'}$. After selecting the label-flipped example $(x', y')$, we add $(x', y')$ to the augmented training set. In other words, we only add an example into the training set if the model $f$ considers the flipped label to be correct. We apply this procedure to each possible label $y' \not= y_i$. In case $S_{i, y'}$ is empty, we do not add any examples to the training set.
% we first construct a set of samples $S_{i,y'}$ with predicted label $y'$ (i.e., $y' = \argmax_y \hat{p}(y | x')$) for all possible labels. If the set $S_{i,y'}$ is empty, we ignore it. Otherwise, we select the sample with the highest probability, i.e.,
% \[
% x', y' = \argmax_{x \in S_{i,y'},y=y'} \hat{p}(y | x)
% \]
% where $x'$ is a sample in the generated set, $y'$ is the flipped label, and the estimated probability $\hat{p}(y' | x')$ scored by the model $f$ is the largest in $S_i$. After selecting the label-flipped example $(x', y')$, we add $(x', y')$ to the augmented training set. In other words, we only add an example into the training set if the model $f$ considers the flipped label to be correct. 
In practice, we find it beneficial to also add the example with the highest probability of label preserving, using the same procedure. After augmenting the training set, we retrain the classifier $f$ to obtain the final model.

% \vspace{-8pt}
\section{Experiments} \label{sec:exp}
% \vspace{-10pt}
% In this section, we conduct extensive experiments on the few-shot version of natural language understanding benchmark SuperGLUE~\cite{wang2019superglue} (also known as FewGLUE~\cite{Schick2021ItsNJ}).
% Results demonstrate that FlipDA is effective in promoting few-shot performance by generating label-flipped data while being robust towards different pretrained models as well as tasks.

\subsection{Experimental Setup}

% \paragraph{Datasets.}
% As is mentioned in Section~\ref{sec:setting}, we use the few-shot SuperGLUE benchmark. 
% Compared to tasks from other NLU benchmarks (e.g., GLUE~\cite{WangSMHLB19}), most of which are single-sentence tasks, SuperGLUE consists of complicated NLU tasks that are all sentence-pair or sentence-triple tasks, which demand more understanding abilities.
% We conduct systematic experiments across 8 SuperGLUE tasks (as described in Section 3.1), and e
% Each task consists of a 32-sample train set, a test set, a validation set, and an additional unlabeled set.
% We conduct systematic experiments across 8 SuperGLUE tasks, ranging from question answering (BoolQ~\cite{Clark2019BoolQET}, MultiRC~\cite{Khashabi2018LookingBT}, and ReCoRD~\cite{Zhang2018ReCoRDBT}), textual entailment (CB~\cite{Marneffe2019TheCI} and RTE~\cite{Dagan2005ThePR}), co-reference resolution (WSC~\cite{Levesque2011TheWS}), causal reasoning (COPA~\cite{Gordon2012SemEval2012T7}), and word sense disambiguation (WiC~\cite{Pilehvar2019WiCTW}).
% Each task consists of a 32-sample train set, a test set, a validation set, and an additional unlabeled set.

\paragraph{Baselines.}
We take seven augmentation methods as the baseline, including Synonym Replacement (SR)~\cite{CharCNN}, KNN Replacement (KNN)~\cite{KNN}, Easy Data Augmentation (EDA)~\cite{EDA},  Back Translation (BT)~\cite{FadaeeBM17a_bt}, TinyBERT (T-BERT)~\cite{TinyBERT}, T5-MLM, and MixUP~\cite{ZhangCDL18_mixup}. For more details about baseline selection and implementation, please refer to Appendix~\ref{apdx:da_baseline}.

\begin{table*}[ht]
\begin{minipage}{\textwidth}
% \begin{table*}[h]
\centering
% \scriptsize
\small
\makeatletter\def\@captype{table}\makeatother\caption{\small{Performance of baseline methods and FlipDA based on PET and ALBERT-xxlarge-v2 (``baseline'' denotes the original PET with no data augmentation. Underline denotes values that outperform ``baseline''. Bold denotes the best-performed ones of the task). ``Avg.'' is the average of scores and ``MD'' (MaxDrop) measures the maximum performance drop over multiple tasks for a given method. All results are the the average over multiple patterns and 3 iterations.}}
% \caption{\small{Performance of baseline methods and FlipDA based on PET and ALBERT-xxlarge-v2 (``baseline'' denotes the original PET with no data augmentation. Underline denotes values that outperform ``baseline''. Bold denotes the best-performed ones of the task). ``Avg.'' is the average of scores and ``MD'' (MaxDrop) measures the maximum performance drop over multiple tasks for a given method. All results are the the average over multiple patterns and 3 iterations.}}
% \vspace{-5pt}
\label{tab:albert}
\begin{tabular}{lcccccccccc}
    % \Xhline{1pt}    
    \toprule[1pt]
     & BoolQ & CB & COPA & RTE & WiC & WSC & MultiRC & ReCoRD \\
    Method & Acc. & Acc./F1 & Acc. & Acc. & Acc. & Acc. & EM/F1a & Acc./F1 & Avg. & MD\\
    % \hline
    \midrule
    Baseline    & 72.47 & 82.74/74.84 & 88.33 & 61.40 & 51.27 & 77.03 & 33.04/74.64 & 86.19/86.75 & 71.20 & -\\
    SR  & \uline{74.98} & \uline{83.33/78.12} & 87.50 & 59.24 & 51.25 & \uline{78.74} & \uline{34.09/75.55} & 85.63/86.12 & \uline{71.64} & 2.16 \\
    KNN & \uline{74.51} & 82.14/74.39 & 85.50 & \uline{61.91} & \uline{51.62} & 75.00 & 32.72/75.20 &84.77/85.31 & 70.73 & 2.83\\
    EDA & \uline{72.68} & 81.10/73.58 & 84.50 & 58.33 & 51.81 & 75.85 & 28.74/73.05 &85.39/85.95 & 69.63 & 3.83 \\
    BT-10 & \uline{74.59} & 82.44/77.72 & 83.00 & 55.93 & 50.77 & 76.82 & 32.96/74.69 & 85.34/85.88 & 70.08 & 5.47\\
    BT-6 & \uline{75.36} & \uline{82.89/76.55} & 86.50 & 57.46 & 51.01 & 77.78 & \uline{34.85/75.82} & 85.83/86.41 & 71.16 & 3.94\\
    T-BERT & \uline{72.60} & \uline{85.42/82.35} & 84.67 & 58.66 & 51.10 & \uline{78.95} & 30.47/73.20 & 84.57/85.12 & 70.82 & 3.66\\
    T5-MLM & \uline{73.86} & \uline{83.48/75.01} & 87.33 & \uline{62.27} & 51.08 & \textbf{79.17} & \uline{33.79/74.06} & 85.15/85.69 & \uline{71.54} & 1.05\\
    MixUP & \uline{75.03} & \uline{83.93/79.28} & 70.33 & \uline{62.06} & \uline{52.32} & 68.70 & \uline{34.06/74.66} & 80.93/81.70 & 68.22 & 18.00\\
    % \hline
    \midrule
    FlipDA & \textbf{76.98} & \textbf{86.31/82.45} & \textbf{89.17} & \textbf{70.67} & \textbf{54.08} & \uline{78.74} & \textbf{36.38/76.23} & \textbf{86.43/86.97} & \textbf{74.63} & \textbf{0.00}\\
    % \Xhline{1pt}
    \bottomrule[1pt]
\end{tabular}
% \vspace{-5pt}
% \end{table*}
\end{minipage}
\vspace{5pt}
\begin{minipage}[t]{\textwidth}
% \begin{table}[h]
% \scriptsize
\small
\makeatletter\def\@captype{table}\makeatother\caption{\small{Performance of baseline methods and FlipDA based on PET and DeBERTa-v2-xxlarge. ``baseline'' denotes the original PET without data augmentation. Underlines denote values that outperform the ``baseline''. `FlipDA cls'' denotes the same classifier as in FlipDA for filtering candidate augmented data. Bold denotes the best-performing ones of the task. Wave-lines denote methods with FlipDA classifiers that outperform the original (without FlipDA classifier) version. ``Avg.'' is the average of scores and ``MD'' (MaxDrop) measures the maximum performance drop over multiple tasks for a given method. All results are the the average over multiple patterns and 3 iterations.}}
% \caption{\small{Performance of baseline methods and FlipDA based on PET and DeBERTa-v2-xxlarge. ``baseline'' denotes the original PET without data augmentation. Underlines denote values that outperform the ``baseline''. `FlipDA cls'' denotes the same classifier as in FlipDA for filtering candidate augmented data. Bold denotes the best-performing ones of the task. Wave-lines denote methods with FlipDA classifiers that outperform the original (without FlipDA classifier) version. ``Avg.'' is the average of scores and ``MD'' (MaxDrop) measures the maximum performance drop over multiple tasks for a given method. All results are the the average over multiple patterns and 3 iterations.}}
% \vspace{-5pt}
\label{tab:deberta}
\centering
    \begin{tabular}{lcccccccccc}
    % \Xhline{1pt}
    \toprule[1pt]
    & BoolQ & CB & COPA & RTE & WiC & WSC & MultiRC & ReCoRD \\
    Method & Acc. & Acc./F1 & Acc. & Acc. & Acc. & Acc. & EM/F1a & Acc./F1 & Avg. & MD\\
    % \hline
    \midrule
    Baseline & 78.30 & 85.42/79.31 & 87.67 & 81.95 & 58.74 & 80.13 & 40.40/78.14 & 90.24/90.77 & 77.36 & -\\
    % \\ \hline
    \midrule
    SR       & 77.37 & \uline{87.20/80.28} & 87.00 & 76.29 & \uline{58.88} & \uline{80.88} & 35.70/76.25 & 89.06/89.55 & 76.18 & 5.66\\ 
    ~+FlipDA cls & \uwave{80.37} & 83.48/79.01 & 85.50 & \uwave{82.79} & \uwave{59.75} & 78.10 & \uwave{37.51/76.84} & \uwave{89.27/89.77} & \uwave{76.81} & \uwave{2.17} \\ \midrule
    KNN     & 75.35 & 83.78/75.61 & 85.00 & 75.45 & \uline{59.63} & 79.38 & 29.84/69.14 & 88.26/88.75 & 74.06 & 9.78  \\
    ~+FlipDA cls & \uwave{78.51} & \uwave{87.50/82.53} & \uwave{88.33} & \uwave{82.79} & 58.66 & 76.39 & \uwave{38.86/77.29} &\uwave{90.31/90.78} & \uline{77.29} & \uline{3.74} \\ \midrule
    EDA     & 74.42 & 83.63/76.23 & 85.83 & 77.38 & 59.28 & 78.74 & 37.02/77.05 & 88.11/88.60 & 75.12 & 4.57\\
    ~+FlipDA cls & \uwave{76.20} & \uwave{87.35/82.35} & \uwave{88.17} & \uwave{82.31} & \uwave{59.94}  & \uwave{79.81} & \uwave{42.84/79.30} & \uwave{90.29/90.77} & \uline{77.86} & \uwave{2.10} \\ \midrule
    BT-10   & 75.38 & \uline{88.24/84.03} & 85.33 & 79.66 & \uline{59.46} & 76.71 & 38.88/77.79 & 90.08/90.56 & 76.42 & 3.42\\
    ~+FlipDA cls & \uwave{79.97} & 85.71/80.50 & \uwave{87.50} & 78.58  & \uwave{60.08} & \uwave{77.24} & \uwave{40.97/78.25} & \uwave{90.39/90.94} & \uwave{77.09} & \uwave{3.37}\\ \midrule
    BT-6    & 76.78 & \uline{86.46/82.56} & 84.00 & 81.47 & 58.69 & 75.11 & \uline{40.53/79.01} &90.20/90.73 & 76.35 & 5.02\\
    ~+FlipDA cls  & \uwave{79.63} & 84.67/77.94  & 77.00 & \uwave{82.91} & \uwave{59.58} & \uwave{77.56} & 39.03/77.64 & \uwave{90.41/90.95} & 75.88 & 10.67\\ \midrule
    T-BERT & 70.53 & \uline{86.01/82.77} & 86.17 & 72.80 & 57.49 & 78.85 & 34.94/75.17 &86.94/87.47 & 74.06 & 9.15\\
    ~+FlipDA cls & \uwave{80.24} & 86.16/81.25 & 83.00 & \uwave{82.19} & \uwave{59.49} & \uwave{79.59} & \uwave{40.78/78.64} & \uwave{90.65/91.17} & \uwave{77.35} & \uwave{4.67} \\ \midrule
    T5-MLM & 77.39 & 83.04/73.71 & \uline{88.17} & 81.23 &\uline{60.73} & \textbf{82.37} & 35.02/74.98 & 89.71/90.25 & 76.66 & 4.27\\
    % \midrule
    MixUP & 63.41 & 71.13/60.83 & 72.00 & 68.59 & 57.70 & 68.38 &39.24/76.88 & 60.12/60.93 & 64.33 & 29.98\\
    \midrule
    FlipDA  & \textbf{81.80} & \textbf{88.24/87.94} & \textbf{90.83} & \textbf{83.75} & \textbf{65.12} & 78.85 & \textbf{44.18/80.00} & \textbf{91.02/91.56} & \textbf{80.23} & \textbf{1.28}\\
    % \Xhline{1pt}
    \bottomrule[1pt]
  \end{tabular}
\vspace{-15pt}
% \end{table}
\end{minipage}
\end{table*}
\paragraph{Evaluation Protocol}
We evaluate augmentation methods based on PET~\cite{Schick2021ItsNJ}.
Following PET, we take a set of pre-fixed hyper-parameters (see Appendix~\ref{apdx:pet_baseline}).
Considering few-shot learning is sensitive to different patterns and random seeds~\cite{DBLP:journals/corr/abs-2002-06305,Schick2021ItsNJ}, we reported the average performance over multiple patterns and 3 iterations.

%Since the instability of few-shot learning~\cite{DBLP:journals/corr/abs-2002-06305} and huge performance difference between patterns \cite{Schick2021ItsNJ}, we reported the average performance over multiple patterns and 3 iterations.

We evaluate FlipDA on 8 tasks with 2 pretrained models.
% To evaluate effectiveness, we use exactly the same metrics (e.g., accuracy, f1 score, and exact match) as PET~\cite{Schick2021ItsNJ}.
For effectiveness, we use exactly the same metrics (i.e., accuracy, F1, and EM) as PET~\cite{Schick2021ItsNJ}.
% PET is a prompt-based training framework, which converts all tasks into cloze problems, and substantially exceeds the previous sequence classification method.
% Additionally, we also explore the robustness of FlipDA with respect to different pretrained models and tasks.
% We experiment on 8 different complicated tasks as mentioned above, and 2 pretrained language models of different scales, respectively ALBERT (ALBERT-xxlarge-v2) and DeBERTa (DeBERTa-xxlarge-v2). 
For robustness, we propose a new metric MaxDrop (MD), which measures the maximum performance drop compared to not using augmentation over multiple tasks for a given method.
% For robustness evaluation, we propose a new metric named MaxDrop (MD), which measures the maximum performance drop compared to not using augmentation over multiple tasks for a given method.
Given tasks $t_1$,...,$t_n$, a target method $M$, and a baseline method $M_B$, MD is defined as
%$$
%\setlength{\abovedisplayskip}{0pt}
% \setlength{\belowdisplayskip}{0pt}
MD=$ 
\max_{t \in \{t_1,...,t_n\}} \max(0, \text{score}_{t,M_B}-\text{score}_{t,M})$, where $\text{score}_{t,M}$ ( $\text{score}_{t,M_B}$) denotes the performance of method $M$ ($M_B$) on task $t$.
% , and $\text{score}_{t,M_B}$ denotes the performance of method $M_B$ on task $t$.
% Smaller values indicate the method is more robust towards various tasks, and vice versa.
Smaller values indicate better robustness w.r.t tasks.
% We follow the same experimental setting as PET/iPET~\cite{Schick2021ItsNJ} where we take a set of fixed hyper-parameters.
% For ALBERT, we use exactly the same hyper-parameters as PET/iPET~\cite{Schick2021ItsNJ}.
% For DeBERTa, we select a set of fixed hyper-params according to practical considerations.
% Please refer to Appendix \ref{apdx:pet_baseline} for details.

\subsection{Main Results}
\begin{table*}[ht]
% \scriptsize
\small
\caption{\small{Ablation study on label-flipped data v.s. label-preserved data on DeBERTa-v2-xxlarge. Bold denotes the best-performed results. Underlines denote the second-best results. ``Avg.'' is the average of scores and ``MD'' (MaxDrop) measures the maximum performance drop over multiple tasks for a given method. All results are the the average over multiple patterns and 3 iterations.}}
\label{tab:flip-preserve}
\centering
  \begin{tabular}{lcccccccc}
    \Xhline{1pt}      
    & BoolQ & CB & COPA & RTE & WiC & MultiRC \\
    Method & Acc. & Acc./F1 & Acc. & Acc. & Acc. & EM/F1a  & Avg. & MD\\
    \hline
    Baseline & 78.30 & 85.42/79.31 & 87.67 & 81.95 & 58.74 & 40.40/78.14 & 74.72 & -\\
    FlipDA (both)  & \textbf{81.80} & \textbf{88.24/87.94} & \textbf{90.83} & \textbf{83.75} & \textbf{65.12} & \textbf{44.18/80.00} & \textbf{78.61} & 0.00 \\
    Label-Flipped & \uline{80.91} & \uline{84.52/80.99} & \uline{89.67} & \uline{83.51} & \uline{62.34} & \uline{42.7/79.37} & \uline{76.70} & 0.00 \\
    Label-Preserved & 77.04 & 83.48/78.68 & 87.67 & 80.99 & 60.08 & 39.55/78.30 & 74.30 & 1.28\\
    \Xhline{1pt}
  \end{tabular}
\end{table*}

\begin{table}%[h]
\small
\caption{\small{Results of different label transformation on DeBERTa.
RTE: A/B denotes entail/not-entail, indicating whether the given premise entails with the given hypothesis.
BoolQ: A/B denotes False/True, representing the answer for the given yes-no questions.
WiC: A/B denotes F/T, indicating whether the target word shares the same meaning in both given sentences.
MultiRC: A/B denotes 0/1, representing whether the given answer is correct for the given question.}}
% RTE: A $\Leftrightarrow$ entail \& B $\Leftrightarrow$ not-entail. 
% BoolQ: A $\Leftrightarrow$ False \& B $\Leftrightarrow$ True. 
% WiC: A $\Leftrightarrow$ F \& B $\Leftrightarrow$ T.
% MultiRC: A $\Leftrightarrow$ 0 \& B $\Leftrightarrow$ 1. }
\label{tab:trans-direction}
\centering
  \begin{tabular}{lcccc}
    % \Xhline{1pt} 
    \toprule[1pt]
    & BoolQ & RTE & WiC & MultiRC \\
    Method & Acc. & Acc. & Acc. & EM/F1a\\
    \midrule
    % baseline & 78.30 & \textbf{81.95} & \textbf{58.74} & 40.40/78.14 \\
    A$\rightarrow$A & 78.89 & 76.17 & 55.66 & 36.57/76.77 \\
    A$\rightarrow$B & 78.34 & \textbf{80.87} & \textbf{57.99} & \textbf{40.94/78.93} \\
    B$\rightarrow$B & 74.55 & 75.57 & 57.30 & 39.73/78.03 \\
    B$\rightarrow$A & \textbf{80.33} & 76.90 & 56.20 & \textbf{40.10/78.41} \\
    % \Xhline{1pt}
    \bottomrule[1pt]
  \end{tabular}
%\vspace{-10pt}
\end{table}
Results are presented in Table~\ref{tab:albert} and Table~\ref{tab:deberta}.
We observe that FlipDA achieves the best performance among all data augmentation methods in both effectiveness (Avg.) and robustness (MD) on both ALBERT-xxlarge-v2 and DeBERTa-v2-xxlarge.

Specifically, FlipDA achieves an average performance of 74.63 on ALBERT-xxlarge-v2 and an average of 80.23 on DeBERTa-v2-xxlarge, both of which outperform baselines by around 3 points.
It suggests FlipDA is effective in boosting the performance of few-shot tasks by augmenting high-quality data without causing too many side effects.

% \begin{table*}[ht]
% % \scriptsize
% \small
% \caption{\small{Ablation study on label-flipped data v.s. label-preserved data on DeBERTa-v2-xxlarge. Bold denotes the best-performed results. Underlines denote the second-best results. ``Avg.'' is the average of scores and ``MD'' (MaxDrop) measures the maximum performance drop over multiple tasks for a given method. All results are the the average over multiple patterns and 3 iterations.}}
% \label{tab:flip-preserve}
% \centering
%   \begin{tabular}{lcccccccc}
%     \Xhline{1pt}      
%     & BoolQ & CB & COPA & RTE & WiC & MultiRC \\
%     Method & Acc. & Acc./F1 & Acc. & Acc. & Acc. & EM/F1a  & Avg. & MD\\
%     \hline
%     Baseline & 78.30 & 85.42/79.31 & 87.67 & 81.95 & 58.74 & 40.40/78.14 & 74.72 & -\\
%     FlipDA (both)  & \textbf{81.80} & \textbf{88.24/87.94} & \textbf{90.83} & \textbf{83.75} & \textbf{65.12} & \textbf{44.18/80.00} & \textbf{78.61} & 0.0 \\
%     Label-Flipped & \uline{80.91} & \uline{84.52/80.99} & \uline{89.67} & \uline{83.51} & \uline{62.34} & \uline{42.7/79.37} & \uline{76.70} & 0.0 \\
%     Label-Preserved & 77.04 & 83.48/78.68 & 87.67 & 80.99 & 60.08 & 39.55/78.30 & 74.30 & 1.28\\
%     \Xhline{1pt}
%   \end{tabular}
% \end{table*}

% We do not implement back-translation on WiC and WSC, because they both need to keep part of the words in the sentence unchanged, which can not be satisfied by back-translation. Meanwhile, we also observe that FlipDA shows improvements on all tasks except WSC, while all the other methods only work on a few tasks (denoted with underlines).
FlipDA shows improvements on all tasks except WSC, while all the other methods only work on a few tasks (denoted with underlines).
Such observations are consistent with the MaxDrop results, where FlipDA achieves the lowest MaxDrop value of 0.0 on ALBERT-xxlarge-v2 and 1.28 on DeBERTa-v2-xxlarge.
This implies FlipDA is robust to different types of tasks, while other augmentation methods could only be effective for partial tasks and not sufficiently robust.

% We also discover two interesting phenomena. The first is that MixUP degrades performance by a lot on the DeBERTa setting. 
% This may be because we fix the bottom 16 layers' parameters to save Video Memory (Appendix \ref{apdx:pet_baseline}), which leads to its poor performance in dealing with the disturbance in the embedding space.
% The second is that our FlipDA in WSC is overtaken by T5-MLM, which is equal to our method without classification. This may be because the predicted result of PET on WSC is 0 or 1, and as a result, the selected examples are not the ones with the highest probability but the first ones with the True label. As a result, our selection method does not improve the accuracy but loses the diversity.
% By comparing the results of albert-xxlarge-v2 and deberta-xxlarge-v2, we can also observe that FlipDA gains even more improvements on the larger-scale DeBERTa.

% T5-MLM, which is equivalent to our method without label flipping, outperforms FlipDA on the WSC task. This may be because 

\subsection{Ablation Study of FlipDA}

% We observe that most of the data augmentation methods show effectiveness in terms of certain tasks or base models to a certain extent, while FlipDA can achieve good performance on almost all the tasks.
% We are especially interested in \textbf{the essential reasons} that make FlipDA effective and robust. 

%In this section, we show ablation results with model DeBERTa-v2-xxlarge. Results with ALBERT are in Appendix~\ref{apdx:albert_result}.

% This section shows ablation results on DeBERTa~\footnote{Results with ALBERT are in Appendix~\ref{apdx:albert_result}}.
% FlipDA generally has two phases, where the first prepares the candidate augmented data and the second selects data. 
% The following experiments study each phase of FlipDA by fixing the other one.
% In the following experiments, we first freeze the second step to study variants the first step, and then fix the first step to study the second.

\paragraph{Effectiveness of Pattern-based Data Cloze}

To study different methods of obtaining candidate augmented data,
% , including replacement-based methods (e.g., KNN replacement and synonym replacement), generation-based methods (e.g., back translation), feature space based methods (e.g., mixup), and our pattern-based data cloze method (i.e., FlipDA), 
we feed candidates obtained by different methods into the same classifier (as FlipDA uses).
% to see if different strategies can also reach similar performance as FlipDA does.
Table~\ref{tab:deberta} shows the ablation results. % on methods of obtaining candidate augmented data. 
% Note that the feature space based method (mixup) is not included since this method can not be combined with a classifier.

FlipDA outperforms all the other baseline methods with a classifier (i.e., with ``FlipDA cls''). 
% FlipDA achieves the largest average effectiveness scores as well as the smallest MaxDrop value (denoted in bold).
% When combining different strategies of obtaining candidate augmented data with FlipDA classifier still cannot reach similar performance to FlipDA, which proves that our pattern-based data cloze strategy with T5 is effective. 
% \zj{I can not understand the grammar, rewrited}
Other methods of obtaining augmented data candidates cannot reach similar performance as FlipDA when combining with FlipDA classifier, which proves the effectiveness of our pattern-based data cloze strategy with T5. 
Reasons could be that T5-based augmentation produces samples with fewer grammatical errors. (will further discuss in Sec \ref{sec:case_study}). Moreover, T5-style blank filling could produce samples that are more compatible with label flipping.

\paragraph{Effectiveness of FlipDA Classifier}

We then compare the performance of different methods with and without the FlipDA classifier.
% \todo{merge table and fix text} From Table~\ref{tab:data-collection-ablation},
% by comparing each baseline method (with FlipDA classifier) with their original version (without FlipDA classifier) in Table~\ref{tab:deberta}, we find that generally, most baseline methods outperform the original version in effectiveness (Avg.) and robustness (MD).
According to Table \ref{tab:deberta}, most baseline methods with the FlipDA classifier outperform the original version in terms of both effectiveness (Avg.) and robustness (MD).
This demonstrates that the FlipDA classifier which is capable of flipping labels and filtering data is effective in augmenting high-quality data and improving few-shot NLU performance.
The only exception is BT-6.
The reason could be data augmented by back translation usually lack diversity and are less likely to change labels, and using the FlipDA classifier further decreases diversity and hurts its performance.
% methods do not change much of the semantics, 
% and thus the obtained augmented data are not diverse enough and are less likely to change labels.
% When back translation is additionally combined with the FlipDA classifier for data filtering, diversity further decreases, resulting in performance degradation compared with the original version.

The improvement brought by the FlipDA classifier is more consistent on BoolQ, RTE, and MultiRC. This may be because these tasks involve predicting a single token with two opposite choices, and thus label flipping might happen more often. Some of the other tasks such as COPA and WSC involve predicting multiple tokens, which makes generating label-flipped data more difficult. This leads to less substantial improvement on these tasks.

\subsection{Analysis of Label-Flipping v.s. Label-Preservation}\label{sec:analysisoflabelflipping}

% Table~\ref{tab:albert} and Table~\ref{tab:deberta} show generating both label-flipped and label-preserved data at the same time lead to performance improvement.
A follow-up question is how label-flipped data and label-preserved data respectively contribute to the overall improvements.
We run decoupling label-flipped data and label-preserved data.
% To answer this question, we conduct further analysis by running decoupling label-flipped data and label-preserved data.
Results are in Table~\ref{tab:flip-preserve}, where bold text represents the best-performed methods.
We conclude that augmenting both label-flipped and label-preserved data leads to the best average performance.
Besides, values with underlines denote the second-best performance, most of which are augmenting only label-flipped data.
Augmenting only label-preserved data leads to the worst performance, even slightly underperforming the non-augmentation baseline.
% When only augmenting label-preserved data, the experiments perform the worst, whose results are even worse than the baseline results. While in ALBERT, the label-preserved data also can still improve the performance. 
This demonstrates the high effectiveness of label-flipping. This aligns well with our analysis in Section \ref{sec:eff}.
% We speculate the reasons could be that label-flipped cases and their corresponding original data examples are much similar while maintaining local subtle differences, which are also the key factors that lead to label-flipping.
% The contrastive information tells more about how to discriminate corresponding task examples.
% More results are in Appendix \ref{apdx:deberta_more} and \ref{apdx:flip-preserve-albert}.
More results on ALBERT are in \ref{apdx:flip-preserve-albert}.

\subsection{Analysis of Label Transformation} \label{sec:trans}

Section~\ref{sec:analysisoflabelflipping} proves that label-flipped augmented data are more effective in improving few-shot performance than label-preserved ones.
% Along this direction, i
It is even more intriguing to study which direction of label flipping is able to benefit the few-shot performance to the maximum extent.
% To account for this, we experiment with 4 tasks, including RTE, BoolQ, WiC, and MultiRC, all of which are binary classification tasks.
% To account for this, 
We experiment with 4 binary classification tasks, i.e., RTE, BoolQ, WiC, and MultiRC.
Each task has 4 directions of label transformation.
We conduct experiments that augment data in each of the four directions respectively and compare their effectiveness. % We take DeBERTa as the base model. Results on ALBERT are in Appendix \ref{apdx:trans-direction-albert}.

\begin{table*}[htbp]
\caption{\small{Results of different strategies for choosing augmented data on DeBERTa (xxlarge). ``Avg.'' is the average of scores and ``MD'' (MaxDrop) measures the maximum performance drop over multiple tasks for a given method. All results are the the average over multiple patterns and 3 iterations.}}
\label{tab:strategies}
\centering
\small
  \begin{tabular}{lcccccccc}
    \Xhline{1pt}      
    & BoolQ & CB & COPA & RTE & WiC & MultiRC \\
    Method & Acc. & Acc./F1 & Acc. & Acc. & Acc. & EM/F1a  & Avg. & MD\\
    \hline
    Baseline & 78.30 & 85.42/79.31 & 87.67 & 81.95 & 58.74 & 40.40/78.14 & 74.72 & -\\
    \hline
    Noisy Student & \textbf{82.13} & 86.31/82.60& 84.33& 82.79 & 64.11 & 39.99/77.43 & 76.09 & 3.34\\
    \hline
    Default Strategy  & {81.80} & {88.24/87.94} & \textbf{90.83} & {83.75} & 65.12 & \textbf{44.18/80.00} & \textbf{78.61} & \textbf{0.00} \\
    Global TopP & 81.22 & 88.10/85.59 & 89.33 & 81.11 & 64.19 & 42.56/79.16 & 77.26 & 0.84\\
    Global TopK & 80.71 & 88.54/85.69 & 87.83 & 81.35 & \textbf{65.13} & 41.14/78.52 & 76.99 & 0.60 \\
    Diverse TopK & \textbf{81.99} & \textbf{89.73/88.92} & 90.0 & \textbf{84.59} & 63.85 & 42.64/79.13 & \textbf{78.44} & \textbf{0.00} \\
    \Xhline{1pt}
  \end{tabular}
\end{table*}

\begin{table*}[h]
  \caption{Some augmented examples selected by our model  (DeBERTa) in RTE. Black denotes original examples, and blue denotes augmented examples.}
%   \vspace{-5pt}
  \label{table:case}
  \small
  \centering
  \begin{tabular}{p{1.5cm}<{\centering}l}
    % \begin{tabular}{c|l}
    % \Xhline{1pt}
    \toprule[1pt]
    \begin{tabular}[c]{p{1.6cm}<{\centering}} Entailment\\ $\rightarrow$ \\ Entailment\end{tabular} & \begin{tabular}[c]{p{0.8\textwidth}}
    % \textbf{Original Example}: \\
    {\begin{tabular}[c]{p{0.8\textwidth}}
    \textbf{Premise:} Tropical Storm Debby is blamed for several deaths across the Caribbean.\\
    \textbf{Hypothesis:} A tropical storm has caused loss of life.\\
    \end{tabular}}
     \\
    % \textbf{Augmented Example}:\\
    {\color[HTML]{3166FF}
    \begin{tabular}[c]{p{0.8\textwidth}}
    \textbf{Premise:} Tropical Storm Maria is blamed for the deaths across the Caribbean\\
    \textbf{Hypothesis:} A hurricane has caused loss of life\\
    \end{tabular}}\\
    \end{tabular} \\
    \midrule
    \begin{tabular}[c]{@{}c@{}}Entailment\\ $\rightarrow$ \\ Not Entailment\end{tabular}                     & \begin{tabular}[c]{p{0.8\textwidth}}
    % \textbf{Original Example}: \\
    {\begin{tabular}[c]{p{0.8\textwidth}}
    \textbf{Premise:} The university server containing the information relating to Mason's ID cards was illegally entered by computer hackers.\\
    \textbf{Hypothesis:} Non-authorized personnel illegally entered into computer networks.\\
    \end{tabular}}
     \\
    % \textbf{Augmented Example}:\\
    {\color[HTML]{3166FF}
    \begin{tabular}[c]{p{0.8\textwidth}}
    \textbf{Premise:} The university server that holds the information about Mason 's ID number was not compromised by hackers\\
    \textbf{Hypothesis:} security personnel illegally hack into computer systems \\
    \end{tabular}}\\
\end{tabular} \\
    \midrule
    \begin{tabular}[c]{@{}c@{}}Not Entailment\\ $\rightarrow$ \\ Entailment\end{tabular}                     & \begin{tabular}[c]{p{0.8\textwidth}}% \textbf{Original Example}: \\
    {\begin{tabular}[c]{p{0.8\textwidth}}
    \textbf{Premise:} Vodafone's share of net new subscribers in Japan has dwindled in recent months.\\
    \textbf{Hypothesis:} There have been many new subscribers to Vodafone in Japan in the past few months.\\
    \end{tabular}}
     \\
    % \textbf{Augmented Example}:\\
    {\color[HTML]{3166FF}
    \begin{tabular}[c]{p{0.8\textwidth}}
    \textbf{Premise:} Vodafone 's number of net new subscribers to Japan has increased in recent months \\
    \textbf{Hypothesis:} There have been net new subscribers to Vodafone in Japan in recent months \\
    \end{tabular}}\\
\end{tabular} \\
\midrule
    \begin{tabular}[c]{@{}c@{}}Not Entailment\\ $\rightarrow$ \\ Not Entailment\end{tabular}                     & \begin{tabular}[c]{p{0.8\textwidth}}
    % \textbf{Original Example}: \\
    {\begin{tabular}[c]{p{0.8\textwidth}}
    \textbf{Premise:} The 10-men team is expected to arrive at the foot of the mountain in the end of April and began their journey to the 8,586-meter peak in early May.\\
    \textbf{Hypothesis:} Kanchenjunga is 8586 meters high.\\
    \end{tabular}}
     \\
    % \textbf{Augmented Example}:\\
    {\color[HTML]{3166FF}
    \begin{tabular}[c]{p{0.8\textwidth}}
    \textbf{Premise:} The 10-men team arrived at the foot of the mountain at the end of March and reached their goal of reaching the 8,586-meter peak in early April\\
    \textbf{Hypothesis:} Kanchenjunga is 8586 meters
    \end{tabular}}\\
\end{tabular} \\
% 	\Xhline{1pt}
\bottomrule[1pt]
% \vspace{-10pt}
  \end{tabular}
\vspace{-5pt}
\end{table*}

% \todo{to be fixed} 
Results on DeBERTa are shown in Table \ref{tab:trans-direction}, and results on ALBERT are in Appendix \ref{apdx:trans-direction-albert}.
% Different tasks can be roughly categorized into two types, asymmetric and symmetric. For BoolQ, RTE, and WiC, transforming in one direction is more beneficial than the other. This demonstrates an effect of asymmetric label transformation. For example, BoolQ aims to answer yes-no questions. It is relatively easy for a model to generate samples with answer ``no'' because one small conflict is sufficient for the prediction. However, in the reverse direction, it is difficult to generate sample with answer ``yes'' as the model has to analyze all details in an entire paragraph to ensure consistency. As a result, ``hard-to-easy'' might benefit from higher generation quality. The above reasoning is also applicable to RTE (i.e., generating non-entailment samples is easier than generating entailment samples) and WiC (i.e., generating sentences in which a word has the same meaning is easier). We observed similar effects for these three tasks that generating ``hard-to-easy'' samples is more beneficial. On the other hand, for MultiRC, the difference between the two directions are not significant. This may be because the two directions are similar in terms of generation difficulty.
% We can see that for all four tasks, at least one label-flipped direction is more beneficial than others, so label-flipping is quite useful. We also observed asymmetric label transformation in BoolQ, RTE, and WiC, which may be due to the augmentation difficulty in different directions.
We can see that some tasks are asymmetric, i.e., transforming in one direction is more beneficial than the other, such as BoolQ, RTE, and WiC. We conjecture that it is because it is relatively easy for a model to generate samples with answers in some direction (from ``yes'' to ``no'' in BoolQ, from 'entailment' to ``not entailment'' in RTE, and so on). While some tasks are symmetric, i.e., the difference between the two directions is not significant, such as MultiRC. 
On all tasks, even though some direction is better than others, augmenting with only one direction will affect the label distribution. This will likely lead to a lower performance than the baseline. Augmenting with all directions is still necessary for the best performance.

\subsection{Analysis of Strategies for Augmented Data Selection}
% \zj{changed this section a lot}

% We also conduct quantitative analysis of the strategies for choosing the augmented data.
% There are several coupled attributes of augmented datasets that would drastically influence the few-shot performance simultaneously, including label distribution, diversity, data quality (since augmented data have noises), etc.
% And unfortunately, we find it hard to decouple each of them to see how each attribute will affect the overall performance.

% We also conduct quantitative analysis of the strategies for choosing the augmented data.
% Several coupled attributes of augmented datasets would drastically influence the few-shot performance simultaneously, including label distribution, diversity, data quality, etc.
% Unfortunately, we find it hard to decouple each of them to see how each attribute will affect the overall performance.

% Therefore, by comprehensively considering the coupled attributes, we propose four plausible strategies for augmented data selection, and quantitatively evaluate them.
% Our proposed four strategies for augmented data selection are described as follows.
% Therefore, by comprehensively considering the coupled attributes, 

We propose four plausible strategies for augmented data selection, and quantitatively evaluate them.
The four strategies are described as follows.
\begin{enumerate}
    \item \textbf{Default Strategy.} It is described in Section \ref{sec:flipda}, with no hyper-parameters. 
    % This is the strategy we described in Section \ref{sec:flipda}.
    % It doesn't require any hyper-parameters.
    % Each sample could have multiple candidate augmented samples. We will divide the candidate augmented samples into different categories by the classifier. For each direction, we will choose the one with the highest probability if there exists. 
    % This method is easy to be implemented and doesn't require any hyper-parameters.
    % \zj{removed the details}
    \item \textbf{Global Top$K$}. 
    % All the candidate augmented data from all the original samples are gathered together and sorted by their predicted probabilities along a certain label transformation direction. The top-$K$ samples with the highest predicted probabilities are selected for the direction. This is equivalently implemented as selected top-$r\%$ augmented samples.
    For each label transformation direction, all the candidate augmented data are gathered and sorted by their predicted probabilities, and the top-$K$ ( or top-$r\%$) samples with the highest probabilities are selected.
    \item \textbf{Global Top$P$}. 
    % Similar to Global TopK, but with a different selection method. Augmented data with predicted probabilities higher than a threshold $P$ are selected.
    Similar to Global Top$K$, but augmented data with predicted probabilities higher than a threshold $P$ are selected.
    % All the candidate augmented data are put together and sorted by their predicted probabilities (along a certain direction). Those whose predicted probabilities are above the preset threshold are selected (as augmented data for the direction).
    % \item \textbf{Diverse TopK}. Each data example is supposed to augment $K$ cases, where $K$ is a predefined hyper-param. For each candidate of each data example, they are first sorted by their predicted probabilities (along certain direction), and the top-K ones are selected as the augmented data in terms of the data example.
    \item \textbf{Diverse Top$K$}. Similar to Global Top$K$ except that a mechanism is used to balance between the original samples. Concretely, we first select the top-1 augmented samples of each original sample (ranked by decreasing probabilities), and then select the top-2, top-3, etc, until $K$ samples have been selected.
    % We will balance the number selected for each example. For each candidate of each data example, they are first sorted by their predicted probabilities (along a certain direction). Then we will first choose the top1 of each example (with probability decreasing), and then the top2, etc.
\end{enumerate}

% \begin{table}[h]
% \small
%   \caption{Different strategies for augmented data selection and their influences to attributes.\XSolidBrush denotes the attribute is changed while \Checkmark means corresponding attribute is kept unchanged.}
%   \label{selection methods}
%   \centering
%   \begin{tabular}{cccc}
%     \hline                 
%     &Label Distribution & Diversity & Data Quality \\
%     \hline
%     Default Strategy & \XSolidBrush &\Checkmark& \XSolidBrush \\
%     Global TopK & \Checkmark & \XSolidBrush & \Checkmark \\
%     Global TopP &\XSolidBrush&\XSolidBrush&\Checkmark \\
%     Diverse TopK &\Checkmark&\Checkmark&\XSolidBrush\\
%     \hline
%   \end{tabular}
% \end{table}

Since FlipDA can be viewed as a self-training algorithm, we also add a self-training algorithm Noisy Student \cite{XieLHL20_noisy_student} as another baseline. 
% We treat the augmented data as unlabeled data and add noise through dropout 0.1. We use spatial dropout in the embedding space. %, or the DeBERTa model will fail.
We treat the augmented data as unlabeled data and add noises with a dropout rate of 0.1.
% through spatial dropout 0.1 in the embedding space. 

Table~\ref{tab:strategies} shows the results of different strategies on different tasks. 
% More results are in Appendix \ref{apdx:deberta_more} and Appendix \ref{apdx:strategies-albert}.
More results on ALBERT are in \ref{apdx:strategies-albert}. 
For Global Top$P$, we set the threshold $P$ at 0.9 or 0.95, whichever is better. For Global Top$K$ and Diverse Top$K$, we select the top 10\% or 20\% augmented examples, whichever is better. 
% Our strategies outperform Noisy Student because as it leverages the idea of label flipping as discussed in Sections \ref{sec:eff} and \ref{sec:robust}. 
Our strategies outperform Noisy Student.
% \zj{removed, not this reason. Because the noisy student could also use the flipped data as unlabeled data. We feed both the preserved data and flipped data into it as unlabeled data to boost its performance. }
Among our four data selection strategies, the Default strategy and Diverse Top$K$ perform the best. Both methods emphasize diversity by using augmented data from different samples. This demonstrates the importance of data diversity and balance for augmented data selection.

\subsection{Case Study} \label{sec:case_study}
% \zj{removed the ``preserved'' cases, and shortened the description}

We show four augmented cases on the RTE task by FlipDA in Table~\ref{table:case}.
Please refer to Appendix~\ref{apdx:case} for more augmented examples.

% We provide two augmented flipped cases on the RTE task by our model in Table \ref{table:case} to show the augmented sample quality of our method. More augmented examples please refer to Appendix~\ref{apdx:case}.
In the first case, we can see that the T5-model changes the name of the tropical storm from ``Debby'' to ``Maria'', and it also changes the ``tropical storm'' to its hypernym ``hurricane'', and all these changes contribute to a different expression without affecting its label. 
The second case adds ``not'' to the premise and therefore the label flips.
The third case changes ``dwindles'' to its antonym ``increased'', and then the label changes from ``Not Entailment'' to ``Entailment''. 
The last case changes the future tense to the simple past tense, ``April'' to ``March'', and ``May'' to ``April'' correspondingly, without affecting its label. 

We can see that the way to change or keep the label is rich and natural. Moreover, the generation quality is improved compared to cases generated by EDA in Table \ref{tab:bad-cases}, which also addresses the concerns of generation quality raised in Section \ref{sec:robust}.

\section{Conclusions}
% \vspace{-5pt}

We propose to study few-shot NLU based on large-scale pretrained models. 
Two key desiderata, i.e., effectiveness and robustness, are identified. 
% Based on the empirical insight that label flipping improves few-shot generalization, we propose FlipDA that utilizes a classifier for automatic label flipping and data selection. 
Based on the empirical insight that label flipping improves few-shot generalization, we propose FlipDA with automatic label flipping and data selection. 
Experiments demonstrate the superiority of FlipDA, outperforming previous methods in terms of both effectiveness and robustness. 
% In the future, it will be crucial to further increase the diversity and quality of augmented data for better performance.
In the future, it will be crucial to theoretically understand why and how generating label-flipped data in the neighborhood of existing data points improves generalization. Moreover, increasing the diversity and quality of augmented data generation is also an important long-term goal.

% \zj{todo: complement the space.}
% theoretically understand why generating label-flipped data in the neighborhood of existing data points improves generalization. Moreover, increasing the diversity and quality of augmented data is also an important long-term goal.

\section*{Acknowledgements}
% Zhou and Li are supported in part by the National Natural Science Foundation of China Grant 61822203, 61772297, 61632016 and Turing AI Institute of Nanjing
% and Xi'an Institute for Interdisciplinary Information Core Technology. 
Zhou and Li are supported in part  by the National Natural Science Foundation of China Grant 62161146004, Turing AI Institute of Nanjing and Xi'an Institute for Interdisciplinary Information Core Technology.
Tang is funded by NSFC for Distinguished Young Scholar (61825602).
Zheng is Funded by China Postdoctoral Science Foundation (2021M690471).
% \end{ack}
% \section*{Acknowledgements}

% This document has been adapted
% by Steven Bethard, Ryan Cotterell and Rui Yan
% from the instructions for earlier ACL and NAACL proceedings, including those for 
% ACL 2019 by Douwe Kiela and Ivan Vuli\'{c},
% NAACL 2019 by Stephanie Lukin and Alla Roskovskaya, 
% ACL 2018 by Shay Cohen, Kevin Gimpel, and Wei Lu, 
% NAACL 2018 by Margaret Mitchell and Stephanie Lukin,
% Bib\TeX{} suggestions for (NA)ACL 2017/2018 from Jason Eisner,
% ACL 2017 by Dan Gildea and Min-Yen Kan, 
% NAACL 2017 by Margaret Mitchell, 
% ACL 2012 by Maggie Li and Michael White, 
% ACL 2010 by Jing-Shin Chang and Philipp Koehn, 
% ACL 2008 by Johanna D. Moore, Simone Teufel, James Allan, and Sadaoki Furui, 
% ACL 2005 by Hwee Tou Ng and Kemal Oflazer, 
% ACL 2002 by Eugene Charniak and Dekang Lin, 
% and earlier ACL and EACL formats written by several people, including
% John Chen, Henry S. Thompson and Donald Walker.
% Additional elements were taken from the formatting instructions of the \emph{International Joint Conference on Artificial Intelligence} and the \emph{Conference on Computer Vision and Pattern Recognition}.

% Entries for the entire Anthology, followed by custom entries

% \newpage

\bibliography{anthology,custom}
\bibliographystyle{acl_natbib}
\clearpage
\appendix

\section{Appendix}
\subsection{More Details about the PET Baseline Implementation}
\label{apdx:pet_baseline}
All experiments are carried out in a Linux environment with a single V100 GPU (32G). 
In order to run each experiment in a single GPU, we fix the bottom 16 layers’ (bottom 1/3 layers) parameters of DeBERTa due to the limitation of GPU memory.

On ALBERT, all the parameters and patterns are kept the same as PET/iPET\cite{Schick2021ItsNJ}. We find that the patterns on RTE give extremely poor results on DeBERTa, so we change the patterns of RTE on DeBERTa for a fair evaluation. Let's denote the hypothesis $h$ and the premise $p$, the new pattern is ``$p \text{Question:}h\text{?Answer:}\_\_\_.$'', while keeping the verbalizer the same as PET/iPET (maps ``entailment'' to ``yes'', ``not entailment'' to ``no''). On DeBERTa, we also reduce the learning rate from 1e-5 to 5e-6 on RTE and WiC, which can improve the baseline a lot. Other settings are kept the same as in ALBERT.

We run each pattern and repetition with seed 42. Different from PET/iPET, to keep the order of the train data loader for different patterns, we will give the train data loader a seed of 10, 20, and 30 for three repetitions.

\subsection{Details of Baseline Augmentation Methods}
We compare our FlipDA with various data augmentation baseline methods. 
We do not choose some generation-based methods \cite{CGBERT_VAE, YooSL19_VAE, LiQTC0Y19_VAE}, because they usually need a lot of training data, which is not suitable for few-shot learning tasks. We also attempted to experiment with methods like LAMBADA \cite{LAMBADA} and GPT3Mix \cite{GPT3Mix}. Because SuperGLUE tasks often involve dependency between sentence pairs, the correlation between augmented sentences is necessary for the data to be meaningful. However, we were not able to generate well-formed, meaningful data from either LAMBADA or GPT3Mix.
% are similar to our work, we find it difficult to apply them on FewGLUE (All examples has two or more sentences). We find them unable to generate desired related sentence pairs.
% First, it is not easy to generate the separator between two sentences automatically. And then, when we force it to generate the separator, we find it is still not easy to generate the related sentences. 
For example, in RTE, we want a premise and a shorter hypothesis that may be contained in the premise, but methods like GPT3Mix usually keep on generating long paragraphs in an uncontrollable manner. Moreover, these methods rely on priming, which is not suitable for datasets with long sentences.

The details of the baseline methods we reported in the paper are as follows.

\label{apdx:da_baseline}
\textbf{Synonym Replacement} (SR)~\cite{CharCNN} augments data by randomly choosing $r\%$ words from original texts (stop words excluded), and replacing them with synonyms from WordNet\footnote{https://wordnet.princeton.edu/}. 
Our implementation is based on parts of the code of EDA \footnote{http://github.com/jasonwei20/eda\_nlp \label{eda_footnote}}. We fix the word replacement ratio to 0.1. We augment 10 times for each sample and then mix them with original samples copied for 10 times. 

\textbf{KNN Replacement} (KNN)~\cite{KNN} is similar with Synonym Replacement but differs in replacing randomly-chosen-words with one of the nearest words derived from GloVe\footnote{https://nlp.stanford.edu/projects/glove/}.
Our implementation is based on parts of the code of TinyBert \footnote{https://github.com/huawei-noah/Pretrained-Language-Model/tree/master/TinyBERT \label{tinybert_footnote}}. We fix the word replacement ratio to 0.1, and replace each word with one of the closest 15 words (K=15) derived from GloVe. We use the word embedding version with 300 dimensions and 6 billion words. We augment 10 times for each sample and then mix them with original samples copied for 10 times. 

\textbf{Easy Data Augmentation} (EDA) \cite{EDA} mixes outputs from four data augmentation methods, including synonym replacement, random insertion, random swap, and random deletion. 
Our implementation is based on the code of EDA \textsuperscript{\ref{eda_footnote}}, which removes all punctuations. Here we implement a new version with punctuation marks since we find them important for hard tasks. All hyper-parameters are kept default, i.e., the four augmentation methods are all with a ratio of 0.1, and each example is augmented 9 times. Finally, we will mix the augmented data with the original data as is done in \cite{EDA}.

\textbf{Back Translation} (BT) \cite{FadaeeBM17a_bt, Sennrich_bt} translates each text into another language, and then back translates into the original language.
% We implement it with the help of Google Translator. 
We implemented two versions of BT with google translator. The first one is BT-10, in which we get the augmented data with 9 languages (Spanish, French, German, Afrikaans, Russian, Czech, Estonian, Haitian Creole, and Bengali) and then mix it with the original sentences. The second one is BT-6, in which we get the augmented data with 5 intermediate languages (Spanish, French, German, Russian, and Haitian Creole) and then mix it with the original sentences.

\textbf{TinyBERT} (T-BERT) \cite{TinyBERT} generates augmented data by randomly (with probability $p$) replacing each token with either word predicted by a Bert-base-cased model (for single-piece word) or words derived by GloVe (for multiple-piece word).
Our implementation is based on the code of TinyBert \textsuperscript{\ref{tinybert_footnote}}. If the sentence length is above 512, we will cut off the sentence. All parameters are kept default. Finally, we mix the augmented data with original examples in equal quantities.

\textbf{T5-MLM}.
We randomly (with probability $p$) mask some tokens, and then fill in the blanks with a large pretrained model. We use pattern-based data cloze to further improve its performance. This is the same as FlipDA with only label-preserved data and without data selection. You can refer to Appendix \ref{apdx:cloze} for more details. We augment with a mask ratio of 0.1 because we find a smaller mask ratio will be better without classification. We augment 10 times for each sample and then mix them with original samples copied for 10 times. 

\textbf{MixUP} \cite{ZhangCDL18_mixup, GuoKR20_mixup} augments data in the feature space, which linearly interpolates between two source sentence embeddings, and correspondingly linearly interpolates the two target embeddings.
For each batch, we first sample $\lambda=Beta(0.5,0.5)$, just as the author \cite{ZhangCDL18_mixup} recommended. Then, we do linear interpolation on the embedding space of two sentences, and make it the input of the model. Finally, we calculate the loss as the interpolation between its outputs and the two targets.

\subsection{Details of Pattern-based Data Cloze Strategy}
\label{apdx:cloze}

Because the target and the format of tasks in FewGLUE vary a lot, adjusting the details of data augmentation for each dataset is necessary. We will always augment in the following three steps: (1) mask the sentence, (2) generate the new label (preserve or flip the label), and (3) fill in the blanks by T5. We also augment 10 times for each example as the candidates. (Augmenting with more times might help, but we only augment 10 times for the sake of time, which has been shown effective in our experiments.)

The T5 model \cite{Raffel2020ExploringTL} is not perfect, especially when it is not finetuned. During our experiments, we find it a good cloze model (good at filling in the blanks with information before or after the blanks) but not a good generation model (not good at generating meaning that is not in the original sentence). As a result, in some tasks whose sentence is short, we induce the T5 model to get some new information by adding extra sentences from other examples in the training data set. 

\textbf{BoolQ}. 
Each example contains two sentences, a question $q$ and a passage $p$. We need to tell whether the answer to the question is True. Let's denote the masked question $masked\_q$ and the masked passage $masked\_p$. If we want to get a True answer, we will feed ``$masked\_q$?Yes, $masked\_p$'' into the model. Otherwise, we will feed ``$masked\_q$?No, $masked\_q$'' into the model. The T5 model will fill the blanks in the masked sentences.

\textbf{CB}.
Each example contains two sentences, a premise $p$ and a hypothesis $h$. We need to tell whether the relationship between the premise and the hypothesis is entailment, contradiction, or neutral.
Let's denote the masked premise $masked\_p$ and the masked hypothesis $masked\_h$. We will feed ``$``masked\_h"~?\_\_\_.~``masked\_p"$'' into the model, where the ``$\_\_\_$'' denotes the mapped answer. Similar to PET, and the verbalizer maps ``entailment'' to ``Yes'',
``contradiction'' to ``No'' and ``neutral'' to ``Maybe''. The T5 model will fill the blanks in the masked sentences, not the answer.

\textbf{COPA}.
Each example contains a premise $p$, a question $q$ (``cause'' or ``effect'') and two choices $c_1$, $c_2$. We need to tell which choice is the cause or effect of the premise.
The sentences in the COPA dataset are much shorter than the others, and the relationship between the three sentences is much more difficult to be represented in one sentence. So we only masked the premise $p$ into $masked\_p$. When we flip the label, we want to make the opposite choice of the candidates ($c_1$ to $c_2$, or $c_2$ to $c_1$), and we also flip the question with a probability of 0.5. If the new question is ``effect'', we will feed  ``$masked\_p$ so that $c_{new\_la}$'' into the model. Otherwise, we will feed ``$masked\_p$, because $c_{new\_la}$'' into the model. Here $new\_la$ denotes the new label.

\textbf{RTE}.
Each example contains two sentences, a premise $p$ and a hypothesis $h$. Our augmentation policy is the same as BoolQ. Let's denote the masked hypothesis $masked\_h$ and the masked premise $masked\_p$. If we want to get a True answer, we will feed ``$masked\_h$?Yes, $masked\_p$'' into the model. Otherwise, we will feed ``$masked\_h$?No, $masked\_q$'' into the model. The T5 model will fill the blanks in the masked sentences.

\textbf{WiC}. 
Each example contains two sentences $s1$ and $s2$, and we need to tell whether the word ``w'' in them has the same meaning. 
If the new label is ``same'', we will feed ``$masked\_s1$. $masked\_s2$. Word `` $w$ '' means the same in the two sentences'' into the model.
Sadly, we find if we concatenate them together with a large mask ratio, they will be similar after filling in the masks. This is because the two sentences are too short and T5 is not ``imaginative'' enough. To solve this problem, if the new label is ``different'', we will augment each sentence separately. We also add one sentence sampled from the training set to urge it to generate a more diverse representation.
We still do not find a perfect way to augment because if a word does not have several meanings, it will be nearly impossible to flip its label from ``same'' to ``different''. We are happy to see that our method can still benefit the model a lot even though it is far from perfect.

\textbf{WSC}.
In our experiments, we find it hard for T5 to generate new entities. In this paper, we do not flip its label, but we do believe that there exists an automatic way to generate good augmented examples with different labels.
\begin{table*}[ht]
\small
\caption{Hyper-parameter search space of our algorithm.}
\label{tab:hyper_space}
\centering
  \begin{tabular}{lccc}
    \toprule[1pt]
    Dataset & Mask Ratio & Fill-in Strategy & Decoding Strategy  \\
    \midrule
    BoolQ & 0.3/0.5 & default & greedy/sample/beam search \\
    CB & 0.5 & default & greedy/sample/beam search\\
    COPA & 0.8 & default/rand\_iter\_1 & greedy/sample/beam search\\
    RTE & 0.5 & default & greedy/sample/beam search\\
    WiC & 0.8 & default & greedy/sample/beam search\\
    WSC & 0.3 & default & greedy/sample/beam search\\
    MultiRC & 0.3/0.5 & rand\_iter\_10 & greedy/sample/beam search\\
    ReCoRD & 0.3 & rand\_iter\_10 &greedy/sample/beam search \\
    \bottomrule[1pt]
\end{tabular}
\end{table*}
\textbf{MultiRC}. 
Each example contains a passage $p$, a question $q$, and several candidate answers $a$s. For each answer, it will have a label $la$. 
Our method is somewhat limited in this task, because it has been ``fliped'' when it is constructed. For the $<p,q,a>$ with label True and $<p,q,a'>$ pair with label False, they have satisfied our key idea: similar but different label examples.
Even though, we still try to flip it more. Let's denote the masked question $masked\_q$, the masked passage $masked\_p$, and masked answer $masked\_a$. We fill feed ``$masked\_q$?~Is~the~correct~answer ``$masked\_a$''?Yes/No.~$masked\_p$'' into the model.

\textbf{ReCoRD}
Each example contains a passage $p$, a question $q$, several candidate entities $e$s, and several possible answers $a$s. We fill first replace the ``@placeholder'' in the question $q$ with new answer $a'$, which is randomly sampled from $e$s in the ``flip'' version and otherwise is sampled from $a$s. Let's denote the masked question $masked\_q$ and the masked passage $masked\_p$. We will feed ``$masked\_q\text{. }masked\_p$'' into the model. Finally, we will substitute the new answer $a'$ in the generated question with ``@placeholder''.

\subsection{Details of Pattern-based Filling-in Strategy}
\label{apdx:fillin}
We conclude three essential factors for the filling-in strategy: the mask ratio, the decoding strategy, and the fill-in strategy.
We divide the mask ratio into three levels: 0.3 (small), 0.5 (medium), and 0.8 (large). The decoding strategy consists of greedy search, random sampling (sample from top 15 words), and beam search (with a beam size of 10). The fill-in strategy consists of filling in the blanks at a time or filling in $k$ blanks at a time iteratively.  Our experiments show that the mask ratio is the key factor.

\subsection{Hyper-parameter Search Space of FlipDA}
\label{apdx:hyperparameter}
We do not search all the possible parameters to save time and avoid overfitting. We are not surprised if there are some better results with a larger search space. Our search space is listed in Table \ref{tab:hyper_space}.

We did preliminary experiments and found some guiding principles. We find that datasets with larger sentence lengths should have a smaller mask ratio, and respectively, datasets with smaller sentence lengths should have a larger mask ratio. (The WSC dataset should be considered separately because we do not flip its label.) We also find that if the sentence length is too large, such as MultiRC or ReCoRD, it is impossible to fill in all the blanks at a time, because the number of blanks may exceed 100. To solve this problem, we fill in 10 random blanks at a time, iteratively until all masks are filled. What's more, the COPA dataset is too short, so we also try to fill in 1 random blank at a time, iteratively until all masks are filled. We do not figure out the relationship between the characteristic of the datasets and the decoding strategies, so we search the three decoding strategies for all datasets. 
% For most of the datasets, greedy or sample is better than beam search. 
For each dataset, we also try two modes: allowing the classifier to change the label or not. (Augmented candidates that are predicted with a different label from the original ones could be dropped.)
Above all, for most of the datasets, we only search 6 hyper-parameter combinations. We think this will not lead to severe overfitting, and our algorithm is stable.
\subsection{Additional Discussion}
\paragraph{Limitations for the WSC Task}
As is illustrated in the body part, label-flipped augmentation has inspiring advantages for few-shot learning performance, but it also has limitations.
While FlipDA significantly outperforms existing baseline augmentation methods on most tasks, we also notice that its effect on the WSC task is a little behind some of the baselines.
This is because, for the WSC task that disambiguates multi-token word senses, it is hard for T5 to generate its label-flipped cases. The T5 model is not good at making up similar entities that are not in the original sentence, and thus unable to produce desired candidate examples.  
We leave a better pattern-based cloze algorithm for such tasks to future work.
We anticipate that entity-centric pretrained models might alleviate this issue \cite{Rosset2020KnowledgeAwareLM}.

\paragraph{Which Few-shot Setting to Use?}
Until now, it still remains an open problem of how to evaluate the performance of few-shot learning.
Currently, there are mainly two mainstream few-shot settings.
The first is to use a set of pre-fixed hyper-parameters that are determined according to practical consideration.
The second is to construct a small dev set (e.g., a 32-sample-dev set), and then perform grid search and use the small dev set for hyper-parameters and model selection.
Our experiments are based on the former setting.
We respectively performed preliminary experiments using both settings and found that the first setting tends to be relatively more stable.
We believe how to evaluate few-shot learning systems is an important research direction for future work, too.

\subsection{More Results on ALBERT}
\label{apdx:albert_result}
In the body part, we only report the ablation results on DeBERTa because the model is larger and seems more stable in our experiments.
In this section, we report ablation results on ALBERT. Most of the conclusions are the same, but some details vary. We conjecture that this might be due to the instability of the training process, the quality of the classification model, or some other unknown issues.

\subsubsection{Effectiveness of Pattern-based Data Cloze and FlipDA Classifier}
\label{apdx:data-collection-ablation}
From Table \ref{tab:data-collection-ablation-albert} we can see that FlipDA is still better than other baselines with a classifier, which means our pattern-based data cloze method will contribute to higher quality data with kept/flipped data.
From the comparison between Table\ref{tab:deberta} and Table \ref{tab:data-collection-ablation-albert}, we can see that the classification is much more useful for DeBERTa than ALBERT. With DeBERTa, almost all augmentation methods will improve their performance with the classifier. With ALBERT, only some augmentation methods will improve its performance on some tasks. This is normal because a better classifier will lead to better classification results, i.e., better-selected augmentation data.

\begin{table*}[!h]
\small
\caption{Ablation study on methods of obtaining candidate augmented data. The ablation study is based on ALBERT-xxlarge-v2. ``cls'' denotes the same classifier as FlipDA for filtering candidate augmented data. Bold denotes the best-performed ones. Wave-lines denotes those that outperforms the original (without FlipDA classifier) version.}
\label{tab:data-collection-ablation-albert}
\centering
  \begin{tabular}{lcccccccc}
    % \Xhline{1pt}  
    \toprule[1pt]
    & BoolQ & CB & COPA & RTE & WiC &  MultiRC \\
    Method & Acc. & Acc./F1 & Acc. & Acc. & Acc. & EM/F1a  & Avg. & MD\\
    \midrule
    Baseline & 72.47 & 82.74/74.84 & 88.33 & 61.40 & 51.27 & 33.04/74.64 & 67.68 &- \\
    SR~+~FlipDA cls & 74.32 & \uwave{84.52/79.32} & 82.17 & \uwave{63.93} & 49.56 & 34.53/74.52 & 67.74 & 6.16 \\
    KNN~+~FlipDA cls & 71.88 & \uwave{84.52/76.83} & 83.17 & \uwave{67.39} & \uwave{53.10} & 31.62/73.92 & \uwave{68.16} & 5.16 \\
    EDA~+~FlipDA cls & \uwave{74.16} & \uwave{84.52/78.92} & 83.00  & \uwave{60.41} & 50.49 & \uwave{34.22/75.52} & \uwave{67.44} & 5.33 \\
    BT-10~+~FlipDA cls & 73.37 & 83.04/74.19 & \uwave{85.00} & \uwave{63.12} & \uwave{51.36} & \uwave{34.60/74.69} & 67.69 & 3.33 \\
    BT-6~+~FlipDA cls  & 73.26 & 80.06/68.59 & \uwave{86.83} & \uwave{61.46} & \uwave{51.72} & 34.49/76.05 & 67.14 & \uwave{4.46} \\
    T-BERT~+~FlipDA cls & \uwave{74.44} & 80.80/73.51 & 84.33 & \uwave{65.40} & 50.19 & \uwave{33.75/74.31} & \uwave{67.59} & 4.00\\
    \hline
    FlipDA  &\textbf{76.98} & \textbf{86.31/82.45} & \textbf{89.17} & \textbf{70.67} & \textbf{54.08} & \textbf{36.38/76.23} & \textbf{71.93} & \textbf{0.00} \\
    \bottomrule[1pt]
\end{tabular}
\end{table*}

\subsubsection{Analysis of Label-Flipping v.s. Label-Preservation}
\label{apdx:flip-preserve-albert}
From Table \ref{tab:flip-preserve-albert}, we can see that FlipDA is still the best, i.e., augmentation with both directions is better than with only one direction. Augmentation with only label-flipped data is better than with only label-preserved data in most tasks. This phenomenon is more obvious with DeBERTa than ALBERT, which may be because the classifier quality of DeBERTa is better than ALBERT. What's more, DeBERTa has learned better representations of similar phrases, so the label-kept examples will contribute less when we experiment with DeBERTa. 
\begin{table*}[!h]
\small
\caption{Ablation study on label-flipped data v.s. label-preserved data on ALBERT-xxlarge-v2. Bold denotes the best-performed results. Underlines denotes the second-best results. ``Avg.'' is the average of scores and ``MD'' (MaxDrop) measures the maximum performance drop over multiple tasks for a given method. All results are the the average over multiple patterns and 3 iterations.}
\label{tab:flip-preserve-albert}
\centering
  \begin{tabular}{lcccccccc}
    \toprule[1pt]     
    & BoolQ & CB & COPA & RTE & WiC & MultiRC \\
    Method & Acc. & Acc./F1 & Acc. & Acc. & Acc. & EM/F1a  & Avg. & MD\\
    \midrule
    Baseline & 72.47 & 82.74/74.84 & 88.33 & 61.40 & 51.27 & 33.04/74.64 & 67.68 & - \\
    FlipDA(both)  &\textbf{76.98} & \textbf{86.31/82.45} & \textbf{89.17} & \textbf{70.67} & \textbf{54.08} & \textbf{36.38/76.23} & \textbf{71.93} & \textbf{0.00} \\
    Label-Flipped & \uline{75.09} & \uline{81.40/73.31} & 86.33 & \uline{67.78} & \uline{53.81} & 32.47/74.67 & 68.99 & 2.00\\
    Label-Preserved & 73.95 & \uline{81.25/74.95} & \uline{87.17} & 64.98 & 51.03 & \uline{34.07/74.81} & 68.27 & 1.16 \\
    \bottomrule[1pt]
  \end{tabular}
\end{table*}

\subsubsection{Analysis of Label Transformation}
\label{apdx:trans-direction-albert}
We took a closer at the effect of label transformation direction in Table \ref{tab:trans-direction-albert}. On BoolQ and RTE, the two flipped directions are better than the kept directions. On all datasets, adding data with more directions is better than with only one direction, even some direction seems extremely bad. This is the same as what we observed with DeBERTa.
\begin{table}[!h]
\small
\caption{Results of different label transformation on ALBERT-xxlarge-v2. 
RTE: A/B denotes entail/not-entail, indicating whether the given premise entails with the given hypothesis.
BoolQ: A/B denotes False/True, representing the answer for the given yes-no questions.
WiC: A/B refers to F/T, indicating whether the target word shares the same meaning in both given sentences.
% RTE: A $\Leftrightarrow$ entail \& B $\Leftrightarrow$ not-entail. 
% BoolQ: A $\Leftrightarrow$ False \& B $\Leftrightarrow$ True. 
% WiC: A $\Leftrightarrow$ F \& B $\Leftrightarrow$ T.}
}
\label{tab:trans-direction-albert}
\centering
  \begin{tabular}{lccc}
    \toprule[1pt]  
    & BoolQ & RTE & WiC  \\
    Method & Acc. & Acc. & Acc.\\
    \midrule
    baseline & 72.47 & 61.40 & 51.27 \\
    A$\rightarrow$A & 71.11 & 63.09 & 51.15 \\
    A$\rightarrow$B & 73.56 & \textbf{66.71} & 51.29 \\
    B$\rightarrow$B & 71.63 & 59.57 & \textbf{52.61} \\
    B$\rightarrow$A & \textbf{74.36} & 65.34 & 49.29 \\
    \bottomrule[1pt]
  \end{tabular}
\end{table}

\subsubsection{Analysis of Strategies for Augmented Data Selection}
\label{apdx:strategies-albert}
From Table \ref{tab:strategies-albert}, we can see that Noisy Student performs well with the ALBERT model. It achieves good results on almost all the datasets except COPA. While with DeBERTa (see Table \ref{tab:strategies}), the Noisy Student is somewhat weaker. This may be because the DeBERTa model fixes the bottom 1/3 layers' parameters to save Video Memory, and thus is not suitable for the perturbation on the embedding space. We have chosen the spatial dropout to alleviate the problem, and it will be much worse with other kinds of dropouts. We think a better self-training policy could further improve the performance of data augmentation.
All other observations of the effectiveness of different strategies are somewhat similar to that with DeBERTa.

\begin{table*}[!h]
\small
\caption{Results of different strategies for choosing augmented data on ALBERT-xxlarge-v2. ``Avg.'' is the average of scores and ``MD'' (MaxDrop) measures the maximum performance drop over multiple tasks for a given method. All results are the the average over multiple patterns and 3 iterations.}
\label{tab:strategies-albert}
\centering
  \begin{tabular}{lcccccccc}
    \toprule[1pt]    
    & BoolQ & CB & COPA & RTE & WiC & MultiRC \\
    Method & Acc. & Acc./F1 & Acc. & Acc. & Acc. & EM/F1a  & Avg. & MD\\
    \midrule
    Baseline & 72.47 & 82.74/74.84 & 88.33 & 61.40 & 51.27 & 33.04/74.64 & 67.68 & - \\
    \midrule
    Noisy Student & \textbf{78.01} & 88.39/83.32 & 82.67 & 69.52 & 54.62 & \textbf{37.02/76.53} & 71.24 & 5.66\\
    \midrule
    Default Strategy  &76.98 & 86.31/82.45 & \textbf{89.17} & \textbf{70.67} & 54.08 & 36.38/76.23 & \textbf{71.93} & \textbf{0.00} \\
    Global TopP & 77.73 & \textbf{88.54/84.88} &87.50 & 67.30 & 54.30 & 35.47/76.47 & 71.59 & 0.83\\
    Global TopK & 76.86 & 87.50/84.42 & 85.33 & 69.43 & 51.97 & 36.48/75.36 & 70.91 & 3.00 \\
    Diverse TopK & 77.27 & 88.39/83.18 & 88.67 & 70.61 & \textbf{55.28} & 32.40/73.64 & 71.77 & 0.82  \\
    \bottomrule[1pt]
  \end{tabular}
\end{table*}

\subsection{Case Study}
\label{apdx:case}
We have provided some flipped augmented examples on RTE in Table \ref{table:case}. Here we provide more augmented examples on other tasks, to be specific, BoolQ, WiC, and COPA. The four datasets cover tasks with different targets and sentence lengths.

\textbf{WiC} is a task to tell whether the word $w$ in the two sentences has the same meaning. From Table \ref{tab:case-wic}, we can see that the two augmented sentences with direction to ``True'' is similar. This is determined by the characteristic of T5. 
% We concatenate them by a prompt and mask it with a large $mask\_ratio$, and as a result, the T5-model will fill in the masks to make every sentence in this paragraph express the same meaning. That's why we augment each sentence separately when the direction is to ``False''. 
In the second case, ``feel'' in ``feel the gravity'' means ``perceive by a physical sensation'', but in ``felt so insignificant'' 
means ``have a feeling or perception about oneself in reaction to someone's behavior or attitude''. The last example violates common sense, but it still can preserve the label and provide diversity, and thus boosting model performance.

\begin{table*}[h]
  \caption{Some augmented examples selected by our model  (DeBERTa) in WiC. Black denotes original examples, and blue denotes augmented examples. Underlines denotes the word to be determined.}
  \label{tab:case-wic}
  \centering
  \begin{tabular}{cl}
    \toprule[1pt]
    \begin{tabular}[c]{@{}c@{}}True\\ $\rightarrow$ \\ True\end{tabular}                     & \begin{tabular}[c]{p{0.8\textwidth}}
    % \textbf{Original Example}: \\
    {\begin{tabular}[c]{p{0.8\textwidth}}
    \textbf{Context 1:} We \uline{vaccinate} against scarlet fever.\\
    \textbf{Context 2:} The nurse \uline{vaccinated} the children in the school.\\
    
    \end{tabular}}
     \\
    % \textbf{Augmented Example}:\\
    {\color[HTML]{3166FF}
    \begin{tabular}[c]{p{0.8\textwidth}}
    \textbf{Context 1:} We \uline{vaccinate} the children against fever and malaria\\
    \textbf{Context 2:} The nurse \uline{vaccinated} the children against fever and malaria\\
    \end{tabular}}\\
    \end{tabular} \\
    \midrule
    \begin{tabular}[c]{@{}c@{}}True\\ $\rightarrow$ \\ False\end{tabular}                     & \begin{tabular}[c]{p{0.8\textwidth}}
    % \textbf{Original Example}: \\
    {\begin{tabular}[c]{p{0.8\textwidth}}
    \textbf{Context 1:} You make me \uline{feel} naked.\\
    \textbf{Context 2:} She \uline{felt} small and insignificant.\\
    \end{tabular}}
     \\
    % \textbf{Augmented Example}:\\
    {\color[HTML]{3166FF}
    \begin{tabular}[c]{p{0.8\textwidth}}
    \textbf{Context 1:} You can \uline{feel} the gravity\\
    \textbf{Context 2:} She \uline{felt} so insignificant and useless \\
    \end{tabular}}\\
\end{tabular} \\
    \midrule
    \begin{tabular}[c]{@{}c@{}}False\\ $\rightarrow$ \\ True\end{tabular}                     & \begin{tabular}[c]{p{0.8\textwidth}}
    % \textbf{Original Example}: \\
    {\begin{tabular}[c]{p{0.8\textwidth}}
    \textbf{Context 1:} Can you \uline{back} up your claims?\\
    \textbf{Context 2:} I can't \uline{back} this plan.\\
    \end{tabular}}
     \\
    % \textbf{Augmented Example}:\\
    {\color[HTML]{3166FF}
    \begin{tabular}[c]{p{0.8\textwidth}}
    \textbf{Context 1:} Can you please \uline{back} to your home \\
    \textbf{Context 2:} I can't \uline{back} from your house \\
    \end{tabular}}\\
\end{tabular} \\
    \midrule
    \begin{tabular}[c]{c}False\\ $\rightarrow$ \\ False\end{tabular}                     & \begin{tabular}[c]{p{0.8\textwidth}}
    % \textbf{Original Example}: \\
    {\begin{tabular}[c]{p{0.8\textwidth}}
    \textbf{Context 1:} Turn and \uline{face} your partner now.\\
    \textbf{Context 2:} The bunkers \uline{faced} north and east, toward Germany.\\
    \end{tabular}}
     \\
    % \textbf{Augmented Example}:\\
    {\color[HTML]{3166FF}
    \begin{tabular}[c]{p{0.8\textwidth}}
    \textbf{Context 1:} Get up and \uline{face} it now\\
    \textbf{Context 2:} The ship \uline{faced} north and south from the coast
    \end{tabular}}\\
\end{tabular} \\
	\bottomrule[1pt]
  \end{tabular}
\end{table*}

\textbf{BoolQ} is a QA task that provides a passage and a question. The author needs to tell whether the answer to the question is True or False according to the given passage. We provide augmented examples of four directions. 
The augmented examples are in Table \ref{tab:case-boolq}. The first case changes ``green onyx'' to ``Brazilian onyx'' without changing its label. The second case changes the passage to make the question True, even though it violates common sense. The third case copies some parts of the passage into the question, and then the label flips. The last case changes the keywords of the example but without changing its label.
\begin{table*}[h]
  \caption{Some augmented examples selected by our model  (DeBERTa) in BoolQ. Black denotes original examples, and blue denotes augmented examples.}
  \label{tab:case-boolq}
  \centering
  \begin{tabular}{cl}
    \toprule[1pt]
    \begin{tabular}[c]{@{}c@{}}True\\ $\rightarrow$ \\ True\end{tabular}                     & \begin{tabular}[c]{p{0.8\textwidth}}
    % \textbf{Original Example}: \\
    {\begin{tabular}[c]{p{0.8\textwidth}}
    \textbf{Passage:} Onyx -- Brazilian green onyx was often used as plinths for art deco sculptures created in the 1920s and 1930s. The German sculptor Ferdinand Preiss used Brazilian green onyx for the base on the majority of his chryselephantine sculptures. Green onyx was also used for trays and pin dishes -- produced mainly in Austria -- often with small bronze animals or figures attached.\\
    \textbf{Question:} is there such a thing as green onyx\\
    
    \end{tabular}}
     \\
    % \textbf{Augmented Example}:\\
    {\color[HTML]{3166FF}
    \begin{tabular}[c]{p{0.8\textwidth}}
    \textbf{Passage:} Onyx is Brazilian Onyx which was often used as the base for art glass sculptures created in the 1920s and 1930s . The German sculptor Ferdinand von Goethe used onyx as the base on the bases of his sculptures . It was also used for making pin plates and pin dishes and many artists produced on-oniex sculptures with various animals and figures attached\\
    \textbf{Question:} Is there such a stone as Brazilian onyx\\
    \end{tabular}}\\
    \end{tabular} \\
    \midrule
    \begin{tabular}[c]{@{}c@{}}True\\ $\rightarrow$ \\ False\end{tabular}                     & \begin{tabular}[c]{p{0.8\textwidth}}
    % \textbf{Original Example}: \\
    {\begin{tabular}[c]{p{0.8\textwidth}}
    \textbf{Passage:} Atomic number -- The atomic number or proton number (symbol Z) of a chemical element is the number of protons found in the nucleus of an atom. It is identical to the charge number of the nucleus. The atomic number uniquely identifies a chemical element. In an uncharged atom, the atomic number is also equal to the number of electrons.\\
% {"question": "is the tick part of arthur's imagination", "passage": "Tick (comics) -- In the Amazon Prime video series, The Tick is fixated on Arthur, and even mentions at one point that his thinking is fuzzy when away from Arthur. Despite Arthur's repeated attempts to push The Tick away, the hero won't leave Arthur's side for long. The Tick also frequently talks about Destiny as if she is a literal person, guiding Arthur's path (``Destiny gave him the suit. I just acted in more of a 'delivery man' role''), alluding to the Parcae in Roman mythology. At one point, Arthur starts to believe that The Tick is merely another hallucination, but that thought is quickly dispelled when Arthur's sister, Dot, interacts with ``The Blue Guy.''''\\
    \textbf{Question:} is the atomic number equal to the number of protons\\
    \end{tabular}}
     \\
    % \textbf{Augmented Example}:\\
    {\color[HTML]{3166FF}
    \begin{tabular}[c]{p{0.8\textwidth}}
    \textbf{Passage:} Atomic number is not equal to atomic number or protons. Atomic number ( A, B, C, Z ) of a chemical element is the number of electrons in the nucleus of an atom . The nucleus is composed by the electrons that are present in the nucleus . The numeric value uniquely identifies an element . In case of uncharged atom , the charge number is equal to the number of electrons\\
    \textbf{Question:} number of the atomic element equal to the number or protons \\
    \end{tabular}}\\
\end{tabular} \\
    \midrule
    \begin{tabular}[c]{@{}c@{}}False\\ $\rightarrow$ \\ True\end{tabular}                     & \begin{tabular}[c]{p{0.8\textwidth}}
    % \textbf{Original Example}: \\
    {\begin{tabular}[c]{p{0.8\textwidth}}
    \textbf{Passage:} Peace bond -- The use of peace bonds is rather uncommon in the U.S. justice system, but a deferred prosecution has a similar effect. Since there is no conviction or admission of any guilt, signing a peace bond in Canada does not usually result in U.S. inadmissibility under INA § 212 (a) (2).\\
    \textbf{Question:} is a peace bond an admission of guilt\\
    \end{tabular}}
     \\
    % \textbf{Augmented Example}:\\
    {\color[HTML]{3166FF}
    \begin{tabular}[c]{p{0.8\textwidth}}
    \textbf{Passage:} Peace bond is an important use of money that is widely used in the U.S. justice system , and deferred prosecution has similar effect . Since there is no promise or admission of guilt in any case , signing a peace bond does not usually result in any conviction under U § 2 ( a ) ( b ) \\
    \textbf{Question:} Is a peace bond part of the criminal justice system \\
    \end{tabular}}\\
\end{tabular} \\
    \midrule
    \begin{tabular}[c]{c}False\\ $\rightarrow$ \\ False\end{tabular}                     & \begin{tabular}[c]{p{0.8\textwidth}}
    % \textbf{Original Example}: \\
    {\begin{tabular}[c]{p{0.8\textwidth}}
    \textbf{Passage:} The Princess and the Goblin (film) -- The Princess and the Goblin (Hungarian: A hercegnő és a kobold) is a 1991 British-Hungarian-American animated musical fantasy film directed by József Gémes and written by Robin Lyons, an adaptation of George MacDonald's 1872 novel of the same name.\\
    \textbf{Question:} is the princess and the goblin a disney movie\\
    \end{tabular}}
     \\
    % \textbf{Augmented Example}:\\
    {\color[HTML]{3166FF}
    \begin{tabular}[c]{p{0.8\textwidth}}
    \textbf{Passage:} The Goblet and the Goblin ( film ) -- The Hound and the Goblin ( Hungarian : A hoz és a kobold ) is a 1996 British-Hungarian-American film directed by Peter Gémes and produced by John Lyons , an adaptation of George MacDonald 's novel of the same name\\
    \textbf{Question:} Is the goblin and the hobbit disney movie
    \end{tabular}}\\
\end{tabular} \\
\bottomrule[1pt]
  \end{tabular}
\end{table*}

\textbf{COPA} is a task that needs to choose the effect or cause of the premise from choice1 and choice2. PET treats it as a multi-token cloze question, i.e., predict the whole sentence of choice1 or choice2. 
% It is easy to get a bad result if we change choice1 or choice2, for it destroys the distribution of language. 
We only change the premise or the question to flip or keep the label.
The augmented examples are in Table \ref{tab:case-copa}. As described in Appendix \ref{apdx:cloze}, there will be three types: keep the label, flip the label but keep the question, and flip the label and the question at the same time. The first case changes `` the archeologist'' to ``she'' and ``site'' to ``earth'', both of them keep the meaning of the sentence. The last three cases change almost the whole sentence, but they are in line with human knowledge.
\begin{table*}[h]
  \caption{Some augmented examples selected by our model  (DeBERTa) in COPA. In this task, we only change the premise or question to flip/keep the label. Black denotes original examples, and blue denotes augmented examples.}
  \label{tab:case-copa}
  \centering
  \begin{tabular}{ccl}
    \toprule[1pt]
    \begin{tabular}[c]{@{}c@{}}Keep-label\end{tabular}
    & \begin{tabular}[c]{@{}c@{}}Keep-question\end{tabular}
    & \begin{tabular}[c]{p{0.6\textwidth}}
    % \textbf{Original Example}: \\
    {\begin{tabular}[c]{p{0.6\textwidth}}
    \textbf{Alternative 1:} She excavated ancient artifacts.\\
    \textbf{Alternative 2:} She read about the site's history.\\
    \\
    \textbf{Premise:} The archeologist dug up the site.\\

    \textbf{Question:} Effect~~~~~~~~\textbf{Correct Alternative:} 0\\
    \end{tabular}}
     \\
    % \textbf{Augmented Example}:\\
    {\color[HTML]{3166FF}
    \begin{tabular}[c]{p{0.6\textwidth}}
    \textbf{Premise:} She dug up the earth.\\
    \textbf{Question:} Effect~~~~~~~~\textbf{Correct Alternative:} 0\\
    \end{tabular}}\\
    \end{tabular} \\
    
    \midrule
    
    % \begin{tabular}[c]{@{}c@{}}Flip-label\end{tabular}
    & \begin{tabular}[c]{@{}c@{}}Keep-question\end{tabular}                  & \begin{tabular}[c]{p{0.6\textwidth}}
    % \textbf{Original Example}: \\
    {\begin{tabular}[c]{p{0.6\textwidth}}
    \textbf{Alternative 1:} She began going to church.\\
    \textbf{Alternative 2:} She began travelling abroad.\\
    \\
    
    \textbf{Premise:} The woman had a religious awakening.\\
    \textbf{Question:} Effect~~~~~~~~\textbf{Correct Alternative:} 0\\
    \end{tabular}}
     \\
    % \textbf{Augmented Example}:\\
    {\color[HTML]{3166FF}
    \begin{tabular}[c]{p{0.6\textwidth}}
    \textbf{Premise:} She had a lot of money.\\
    \textbf{Question:} Effect~~~~~~~~\textbf{Correct Alternative:} 1\\
    \end{tabular}}\\
    \end{tabular} \\
    \cline{2-3}

    \begin{tabular}[c]{@{}c@{}}Flip-label\end{tabular} &\begin{tabular}[c]{@{}c@{}}Flip-question\\\\ (Effect \\ $\rightarrow$ \\ Cause)\end{tabular}
    & \begin{tabular}[c]{p{0.6\textwidth}}
    % \textbf{Original Example}: \\
    {\begin{tabular}[c]{p{0.6\textwidth}}
    \textbf{Alternative 1:} Her friend sent her a greeting card.\\
    \textbf{Alternative 2:} Her friend cut off contact with her.\\
    \\
    
    \textbf{Premise:} The woman betrayed her friend.\\
    \textbf{Question:} Effect~~~~~~~~\textbf{Correct Alternative:} 1\\
    \end{tabular}}
     \\
    % \textbf{Augmented Example}:\\
    {\color[HTML]{3166FF}
    \begin{tabular}[c]{p{0.6\textwidth}}
    \textbf{Premise:} A woman is happy.\\
    \textbf{Question:} Cause~~~~~~~~\textbf{Correct Alternative:} 0\\
    \end{tabular}}\\
    \end{tabular} \\
    \cline{3-3}
    
    % \begin{tabular}[c]{@{}c@{}}Flip-label\end{tabular} 
    & \begin{tabular}[c]{@{}c@{}}Flip-question\\\\ (Cause \\ $\rightarrow$ \\ Effect) \end{tabular}
    & \begin{tabular}[c]{p{0.6\textwidth}}
    % \textbf{Original Example}: \\
    {\begin{tabular}[c]{p{0.6\textwidth}}
    \textbf{Alternative 1:} The cafe reopened in a new location.\\
    \textbf{Alternative 2:} They wanted to catch up with each other.\\
    \\
    
    \textbf{Premise:} The women met for coffee.\\
    \textbf{Question:} Cause~~~~~~~~\textbf{Correct Alternative:} 1\\
    \end{tabular}}
     \\
    % \textbf{Augmented Example}:\\
    {\color[HTML]{3166FF}
    \begin{tabular}[c]{p{0.6\textwidth}}
    \textbf{Premise:} The cafe closed.\\
    \textbf{Question:} Effect~~~~~~~~\textbf{Correct Alternative:} 0\\
    \end{tabular}}\\
    \end{tabular} \\
	\bottomrule[1pt]
  \end{tabular}
\end{table*}

\end{document}